\documentclass{tlp}

\usepackage[T1]{fontenc}
\usepackage{amssymb}
\usepackage{amsmath}
\usepackage{epsfig}
\usepackage{theorem}
\usepackage{url}

\DeclareMathSymbol{\naf}{\mathord}{symbols}{"18}

\newcommand{\nat}{\mathbb{N}}
\newcommand{\bbox}{\blacksquare}
\newcommand{\union}{\cup}
\newcommand{\isect}{\cap}
\newcommand{\join}{\sqcup}
\newcommand{\sjoin}{\underline{\sqcup}}
\newcommand{\lpeq}[1]{\equiv_{\mathrm{#1}}}

\newcommand{\fun}[2]{:{#1}\rightarrow{#2}}
\newcommand{\tuple}[1]{\langle{#1}\rangle}
\newcommand{\pair}[2]{\langle{#1},{#2}\rangle}
\newcommand{\rg}[3]{{#1}{#2}\ldots{#2}{#3}}
\newcommand{\set}[1]{\{#1\}}
\newcommand{\sel}[2]{\{{#1}\mid{#2}\}}
\newcommand{\eset}[2]{\{{#1},\ldots,{#2}\}}
\newcommand{\dep}[1]{\mathrm{Dep}^+\!(#1)}

\newcommand{\module}[1]{\mathbb{#1}}
\newcommand{\system}[1]{\textsc{#1}}

\newcommand{\GLred}[3]{{#1}^{#2,#3}}
\newcommand{\sm}[1]{\mathrm{SM}(#1)}
\newcommand{\cm}[1]{\mathrm{CM}(#1)}
\newcommand{\lm}[1]{\mathrm{LM}(#1)}

\newcommand{\IF}{\leftarrow}
\newcommand{\rsep}{.\;}
\newcommand{\choice}[1]{\{#1\}}
\newcommand{\limit}[2]{{#1}\leq\{{#2}\}}

\newcommand{\hb}[1]{\mathrm{At}(#1)}
\newcommand{\hbv}[1]{\mathrm{At_v}(#1)}
\newcommand{\hbh}[1]{\mathrm{At_h}(#1)}
\newcommand{\hbi}[1]{\mathrm{At_i}(#1)}
\newcommand{\hbo}[1]{\mathrm{At_o}(#1)}

\newcommand{\hid}[1]{#1_{\mathrm{h}}}

\newcommand{\head}[1]{\mathrm{Head}(#1)}
\newcommand{\pbody}[1]{\mathrm{Body}^+\!(#1)}
\newcommand{\nbody}[1]{\mathrm{Body}^-\!(#1)}
\newcommand{\body}[1]{\mathrm{Body}(#1)}
\newcommand{\choiceheads}[1]{\mathrm{Choices}(#1)}

\newcommand{\bottom}[2]{\mathrm{b}_{#2}(#1)}
\newcommand{\rest}[3]{\mathrm{e}(\topp{#1}{#2},#3)}
\newcommand{\topp}[2]{\mathrm{t}_{#2}(#1)}

\newcommand{\tr}[2]{\mathrm{Tr}_{\mathrm{#1}}(#2)}
\newcommand{\trop}[1]{\mathrm{Tr}_{\mathrm{#1}}}
\newcommand{\eqt}{\mathrm{EQT}}

\newcommand{\clark}[1]{\mathrm{Comp}(#1)}
\newcommand{\loops}[2]{\mathrm{LF}(#1,#2)}
\newcommand{\lfs}[1]{\mathrm{LF}(#1)}

\newcommand{\reveal}[1]{\mathrm{reveal}(#1)}

\makeatletter
\def\@underjournal{%
  \vbox to 5.6\p@{\noindent\parbox[t]{4.8in}{\normalfont\affilsize\rmfamily
    {\itshape To appear in
     Theory and Practice of Logic Programming\/}\\[2.5\p@]
     \ }%
  \vss}%
}
\let\@j@urnal\@underjournal
\makeatother

\title[Theory and Practice of Logic Programming]
{Achieving Compositionality of \\
the Stable Model Semantics \\
 for \system{smodels} Programs
\thanks{%
This is an extended version of two conference papers
\protect\cite{OJ06:ecai,Oikarinen07:lpnmr}
presented at ECAI'06 and LPNMR'07, respectively.}}

\author[E. Oikarinen and T. Janhunen]
{EMILIA OIKARINEN and TOMI JANHUNEN \\
 Helsinki University of Technology TKK \\
 Department of Information and Computer Science \\
 P.O. Box 5400, FI-02015 TKK, Finland \\
 \url{Emilia.Oikarinen@tkk.fi, Tomi.Janhunen@tkk.fi}}

\makeatletter
\def\@maketitle#1{%
  \newpage
  \vspace*{-15\p@}%
  {\centering \sloppy
   \pe@rl{#1}%
   {\normalfont\LARGE\itshape \@title\par}%
   \vskip 16\p@
   {\normalfont\normalsize\rmfamily
    \let\authorbreak\auth@rbreak
    \let\and\@nd
    \b@at
      \@author
    \end{author@tabular}%
   \par}%
     \vskip 10pt%
     {{\affilsize\it submitted \@submitted; accepted \@accepted}}\par
  }%
  \vskip 18\p@ \@plus 2\p@ \@minus 1\p@
}
\makeatother
\submitted{29 February 2008}
 \accepted{25 September 2008}

\newtheorem{theorem}{Theorem}[section] 
\newtheorem{corollary}[theorem]{Corollary} 
\newtheorem{definition}[theorem]{Definition} 
 
\newtheorem{lemma}[theorem]{Lemma} 
\newtheorem{example}[theorem]{Example} 

\begin{document}
\maketitle

\begin{abstract}
In this paper, a Gaifman-Shapiro-style module architecture is
tailored to the case of \system{smodels} programs under the
stable model semantics. The composition of \system{smodels} program
modules is suitably limited by module conditions which ensure the
compatibility of the module system with stable models. Hence
the semantics of an entire \system{smodels} program depends
directly on stable models assigned to its modules.
This result is formalized as a \emph{module theorem} which truly
strengthens Lifschitz and Turner's splitting-set theorem
\citeyear{LT94:iclp} for the class of \system{smodels} programs.
To streamline generalizations in the future, the module theorem is
first proved for normal programs and then extended to cover
\system{smodels} programs using a translation from the latter
class of programs to the former class.
Moreover, the respective notion of module-level equivalence, namely
\emph{modular equivalence}, is shown to be a proper congruence
relation: it is preserved under substitutions of modules that
are modularly equivalent.
Principles for program decomposition are also addressed. The strongly
connected components of the respective dependency graph can be
exploited in order to extract a module structure when there is no
explicit a priori knowledge about the modules of a program. The paper
includes a practical demonstration of tools that have been developed
for automated (de)composition of \system{smodels} programs.
\end{abstract}

\begin{keywords}
answer set programming, module system, 
compositional semantics,
stable model semantics, 
modular equivalence
\end{keywords}

%------------------------------------------------------------------------------
\section{Introduction}

\emph{Answer set programming} (ASP)
\cite{Niemela99:amai,MT99:slp,GL02:aij,Baral:knowledge}
is an approach to declarative rule-based constraint programming that
has been successively used in many knowledge representation and
reasoning tasks
\cite{DBLP:conf/asp/SoininenNTS01,%
DBLP:conf/padl/NogueiraBGWB01,%
DBLP:journals/tplp/ErdemLR06,Brooks:inferring}.
In ASP, the problem at hand is solved declaratively
\begin{enumerate}
\item
by writing down a logic program the answer sets of which correspond
to the solutions of the problem and

\item
by computing the answer sets of the program using a special purpose
search engine that has been designed for this task.
\end{enumerate}
A modelling philosophy of this kind suggests to treat programs as
integral entities. The answer set semantics---originally defined for
entire programs only~\cite{GL88:iclp,GL90:iclp}---reflects also this
fact. 
Such indivisibility of programs is creating an increasing problem as
program instances tend to grow along the demands of new application
areas of ASP. It is to be expected that prospective application areas
such as semantic web, bioinformatics, and logical cryptanalysis will
provide us with huge program instances to create, to solve, and to
maintain.

Modern programming languages provide means to exploit
\emph{modularity} in a number of ways to govern the complexity of
programs and their development process. Indeed, the use of
\emph{program modules} or \emph{objects} of some kind can be viewed as
an embodiment of the classical \emph{divide-and-conquer} principle in
the art of programming.
The benefits of modular program development are numerous. A software
system is much easier to design as a set of interacting components
rather than a monolithic system with unclear internal structure.  A
modular design lends itself better for implementation as programming
tasks are then easier to delegate amongst the team of programmers. It
also enables the re-use of code organized as module libraries, for
instance.
To achieve similar advantages in ASP, one of the central goals of
our research is to foster modularity in the context of ASP.

Although modularity has been studied extensively in the context of
conventional logic programs, see Bugliesi et al.~\shortcite{BLM94} for
a survey, relatively little attention has been paid to modularity in
ASP. Many of the approaches proposed so far are based on very strict
syntactic conditions on the module hierarchy, for instance, by enforcing
stratification of some kind, or by prohibiting recursion altogether
\cite{EGV97,TBA05:asp,EGM97:acm}.
On the other hand, approaches based on \emph{splitting-sets}
\cite{LT94:iclp,EGV97,FGL05:icdt}
are satisfactory from the point of view of
\emph{compositional semantics}:
the answer sets of an entire program are obtained as specific
combinations of the answer sets of its components. A limitation
of splitting-sets is that they divide logic programs in two
parts, the \emph{top} and the \emph{bottom}, which is rather
restrictive and conceptually very close to stratification.
On the other hand, the compositionality of answer set semantics is
neglected altogether in syntactic approaches
\cite{IIPSC04:nmr,TBA05:asp} and this aspect of models remains
completely at the programmer's responsibility.

To address the deficiencies described above, we accommodate a module
architecture proposed by Gaifman-Shapiro~\shortcite{GS89} to answer
set programming, and in particular, in the context of the \system{smodels}
system \cite{SNS02:aij}.
\footnote{Also other systems such as \system{clasp} \cite{GKNS07:lpnmr}
and \system{cmodels} \cite{Lierler05:lpnmr} that are compatible with
the internal file format of the \system{smodels} system are implicitly
covered.}
There are two main criteria for the design. First of all, it is
essential to establish the full compositionality of answer set
semantics with respect to the module system. This is to ensure
that various reasoning tasks---such as the verification of
program equivalence \cite{JO07:tplp}---can be modularized.
Second, for the sake of flexibility of knowledge representation, any
restrictions on the module hierarchy should be avoided as far as
possible.
We pursue these goals according to the following plan.

In Section~\ref{section:related}, we take a closer look at modularity
in the context of logic programming. In order to enable comparisons
later on, we also describe related approaches in the area of ASP in
more detail.
The technical preliminaries of the paper begin with a recapitulation of
\emph{stable model semantics}~\cite{GL88:iclp}
in Section~\ref{section:background}. However, stable models, or answer
sets, are reformulated for a class of programs that corresponds to the
input language of the \system{smodels} solver~\cite{SNS02:aij}. The
definition of \emph{splitting-sets} is included to enable a detailed
comparison with our results. Moreover, we introduce the concepts of
\emph{program completion} and \emph{loop formulas}
to be exploited in proofs later on, and review some notions of
equivalence that have been proposed in the literature.

In Section \ref{section:modules}, we present a \emph{module architecture}
for \system{smodels} programs, in which the interaction between
modules takes place through a clearly defined
\textit{in\-put/out\-put interface}.
The design superficially resembles that of Gaifman and
Shapiro~\shortcite{GS89} but in order to achieve the full
compositionality of stable models, further conditions on program
composition are incorporated. This is formalized as the main result of
the paper, namely the \emph{module theorem}, which goes beyond the
splitting-set theorem~\cite{LT94:iclp} as \textit{negative recursion}
is tolerated by our definitions. The proof is first presented for
normal programs and then extended for \system{smodels} programs using
a translation-based scheme. The scheme is based on three distinguished
properties of translations, \emph{strong faithfulness},
\emph{preservation of compositions}, and \emph{modularity}, that are
sufficient to lift the module theorem. In this way, we get prepared for
even further syntactic extensions of the module theorem in the future.
The respective notion of module-level equivalence, that is,
\emph{modular equivalence}, is proved to be a proper \emph{congruence}
for program composition. In other words, substitutions of modularly
equivalent modules preserve modular equivalence.
This way modular equivalence can be viewed as a reasonable
compromise between \emph{uniform equivalence}~\cite{EF03:iclp} which
is not a congruence for program union, and \emph{strong
  equivalence}~\cite{LPV01:acmtocl}
which is a congruence for program union but allows only
rather straightforward semantics-preserving transformations of (sets
of) rules. 

In Section \ref{section:decomposition-and-semantical-join}, we address
principles for the decomposition of \system{smodels} programs. It
turns out that strongly connected components of dependency graphs
can be exploited in order to extract a module structure when there is
no explicit a priori knowledge about the modules of a program.  In
addition, we consider the possibility of relaxing our restrictions on
program composition using the content of the module theorem as a
criterion.  The result is that the notion of modular equivalence
remains unchanged but the computational cost of checking legal
compositions of modules becomes essentially higher.
In Section \ref{section:experiments}, we demonstrate how the module
system can be exploited in practise in the context of the
\system{smodels} system.  
We present tools that have been developed for (de)composition of
logic programs and conduct a practical experiment which illustrates
the performance of the tools when processing very large benchmark
instances, that is, \system{smodels} programs having up to millions of
rules. 
The concluding remarks of this paper are presented in Section
\ref{section:conclusions}.  

%------------------------------------------------------------------------------

\section{Modularity aspects of logic programming}
\label{section:related}

Bugliesi et al.~\shortcite{BLM94} address several properties that are
expected from a modular logic programming language. For instance, a
modular language should 
\begin{itemize}
\item
allow {\em abstraction, parameterization}, and {\em information hiding}, 
\item
{\em ease program development} and {\em maintenance} of large programs, 
\item
allow {\em re-usability}, 
\item
have a {\em non-trivial notion of program equivalence} to
justify replacement of program components, and 
\item
maintain the {\em declarativity} of logic programming. 
\end{itemize}
Two mainstream programming disciplines are identified. 
In {\em pro\-gram\-ming-in-the-large} approaches programs are composed
with algebraic operators, see for instance
\cite{Keefe85,MP88:iclp,GS89,BMPT94}.
In {\em programming-in-the-small} approaches abstraction mechanisms
are used, see for instance \cite{Miller86,GM94:jlp}.    

The {\em programming-in-the-large} approaches have their roots
in the framework proposed by O'Keefe~\shortcite{Keefe85} where logic
programs are seen as {\em  elements of an algebra} and the
operators  for {\em composing programs} are seen as {\em operators in
  that algebra}.  
The fundamental idea is that a logic program should be understood
as a part of a system of programs.
Program composition is a powerful tool for structuring programs
without any need to extend the underlying language of Horn clauses. 
Several algebraic operations such as {\em union}, {\em deletion}, {\em
overriding union} and {\em closure} have been considered. 
This approach {\em supports} naturally the {\em re-use} of the pieces 
of programs in different composite programs, and when combined with an
adequate equivalence relation also the replacement of equivalent
components.  
This approach is highly flexible, as new composition mechanisms can
be obtained by introducing a corresponding operator in the algebra or
combining existing ones. 
Encapsulation and information hiding can be obtained by
introducing suitable {\em interfaces} for components.

The {\em programming-in-the-small} approaches originate from
\cite{Miller86}. In this approach the composition of modules is 
modelled in terms of logical connectives of a language that is defined
as an {\em extension of Horn clause logic}. The approach
in~\cite{GM94:jlp} employs the same structural properties, but
suggests a more refined way of modelling visibility rules than the one
in~\cite{Miller86}. 

It is essential that a semantical characterization of a modular
language is such that the meaning of composite programs can be defined
in terms of the meaning of its components~\cite{Maher93}. To be able
to identify when it is safe to substitute a module with another
without affecting the global behavior it is crucial to have a notion
of {\em semantical equivalence}.  
More formally these desired properties can be described under the
terms of {\em compositionality} and {\em full
abstraction}~\cite{GS89,Meyer88}.   

Two programs are {\em observationally congruent}, if and only if they
exhibit the same observational behavior in every {\em context} they
can be placed in.  
A semantics is compositional if semantical equality implies
observational congruence, and
fully abstract if semantical equivalence coincides with 
observational congruence. 
The compositionality and full abstraction properties for different
notions of semantical equivalence ({\em subsumption equivalence}, {\em
logical equivalence}, and {\em minimal Herbrand model equivalence})
and different operators in an algebra (union, closure, overriding
union) are considered in~\cite{BLM94}. 
It is worth noting that minimal Herbrand model equivalence
coincides with the \emph{weak equivalence} relation for positive logic
programs.
As to be defined in Section \ref{sect:equivalences}, two logic
programs are weakly equivalent if and only if they have exactly the
same answer sets. 
As the equivalence based on minimal Herbrand model semantics is not
compositional with respect to program union~\cite{BLM94}, 
we note that it is not a suitable composition operator for
our purposes unless further constraints are introduced. 

\subsection{Modularity in answer set programming}

There are a number of approaches within answer set programming
involving modularity in some sense, but only a few of them really
describe a flexible module architecture with a clearly defined
interface for module interaction.

Eiter, Gottlob, and Veith~\shortcite{EGV97} address modularity in 
ASP in the programming-in-the-small sense.
They view program modules as {\em generalized
  quantifiers} as introduced in~\cite{Mostowski57}.
The definitions of quantifiers are allowed to nest, that is, program
$P$ can refer to another module $Q$ by using it as a generalized
quantifier.  
The main program is clearly distinguished from subprograms, and it is
possible to nest calls to submodules if the so-called {\em call graph}
is {\em hierarchical}, that is, {\em acyclic}. Nesting, however,
increases the computational complexity depending on the depth of nesting.  

Ianni et al.~\shortcite{IIPSC04:nmr} propose another
programming-in-the-small approach to ASP based on {\em
  templates}. 
The semantics of programs containing template atoms is determined by
an \emph{explosion algorithm}, which basically replaces the template
with a standard logic program. 
However, the explosion algorithm is not guaranteed to terminate if
template definitions are used recursively.

Tari et al.~\shortcite{TBA05:asp} extend the language of normal
logic programs by introducing the concept of {\em import rules} for
their ASP program modules.
There are three types of import rules which are used to import a set of
tuples $\overline{X}$ for a predicate $q$ from another module.  
An {\em ASP module} is defined as a quadruple of a module name, a set
of parameters, a collection of normal rules and a collection of import
rules.   
Semantics is only defined for modular programs with acyclic
dependency graph, and answer sets of a module are defined with respect
to the modular ASP program containing it. Also, it is required that
import rules referring to the same module always have the same form. 

Programming-in-the-large approaches in ASP are mostly
based on Lifschitz and Turner's splitting-set theorem \cite{LT94:iclp}
or are variants of it. 
The class of logic programs considered in \cite{LT94:iclp} is that of
{\em extended disjunctive logic programs}, that is, disjunctive logic
programs with two kinds of negation.
A {\em component structure} induced by a {\em splitting sequence},
that is, iterated splittings of a program, allows a bottom-up
computation of answer sets.
The restriction implied by this construction is that the dependency
graph of the component chain needs to be acyclic.

\newcommand{\sche}[1]{\mathbf{#1}}

Eiter, Gottlob, and Mannila \shortcite{EGM97:acm} consider {\em
disjunctive logic programs as a query language for relational
databases}.
A query program $\pi$ is instantiated with respect to an input
database $D$ confined by an input schema $\sche{R}$. The semantics of
$\pi$ determines, for example, the answer sets of $\pi[D]$ which are 
projected with respect to an output schema $\sche{S}$.
Their module architecture is based on both {\em positive and negative
  dependencies} and no recursion between modules is tolerated. 
These constraints enable a straightforward generalization of the
splitting-set theorem for the architecture. 

Faber et al.~\shortcite{FGL05:icdt} apply the {\em magic set
method} in the evaluation of {\em Datalog programs with negation},
that is, effectively normal logic programs. This involves the concept
of an {\em independent set} $S$ of a program $P$ which is a
specialization of a splitting set. 
Due to a close relationship with splitting sets, the flexibility of
independent sets for parceling programs is limited in the same way.   

The approach based on {\em lp-functions}~\cite{GG99,Baral:knowledge}
is another programming-in-the-large approach.
An lp-function has an interface based on input and output signatures. 
Several operations, for instance {\em incremental extension}, {\em
  interpolation}, {\em input opening}, and {\em input extension}, are
introduced for composing and refining lp-functions.
The composition of lp-functions, however, only allows incremental
extension, and thus similarly to the splitting-set theorem there can
be no recursion between lp-functions.

\renewcommand{\GLred}[2]{{#1}^{#2}}

%------------------------------------------------------------------------------

\section{Preliminaries: {\sc smodels} programs}
\label{section:background}

To keep the presentation of our module architecture compatible with an
actual implementation, we cover the input language of the
\system{smodels} system---excluding {\em optimization statements}.  
In this section we introduce the syntax and semantics for
\system{smodels} programs, and, in addition, point out a number of
useful properties  of logic programs under stable model semantics. We
end this section with a review of equivalence relations that have been
proposed for logic programs. 

\subsection{Syntax and semantics}

\emph{Basic constraint rules} \cite{SNS02:aij} are either
\emph{weight rules} of the form
\begin{equation}
\label{eq:weight-rule}
a\IF\limit{w}{\rg{b_1=w_{b_1}}{,}{b_n=w_{b_n}},
              \rg{\naf c_1=w_{c_1}}{,}{\naf c_m=w_{c_m}}}
\end{equation}
or {\em choice rules} of the form
\begin{equation}
\label{eq:choice-rule}
\choice{\rg{a_1}{,}{a_h}}\IF\rg{b_1}{,}{b_n},\rg{\naf c_1}{,}{\naf c_m}
\end{equation}
where $a$, $a_i$'s, $b_j$'s, and $c_k$'s are atoms, $h>0$, $n\geq 0$,
$m\geq 0$, and $\naf$ denotes \emph{negation as failure} or
\emph{default negation}. In addition, a weight rule
(\ref{eq:weight-rule}) involves a weight limit $w\in\nat$ and the
respective weights $w_{b_j}\in\nat$ and $w_{c_k}\in\nat$ associated
with each \emph{positive literal} $b_j$ and \emph{negative literal}
$\naf c_k$.
We use a shorthand $\naf A=\set{\naf a\mid a\in A}$ for any set of
atoms $A$.
Each basic constraint rule $r$ consists of two parts: $a$ or
$\eset{a_1}{a_h}$ is the \emph{head} of the rule, denoted by
$\head{r}$, whereas the rest is called its \emph{body}.
The set of atoms appearing in a body of a rule can be further divided
into the set of {\em positive body atoms},  
defined as $\pbody{r}=\set{b_1,\ldots, b_n}$, and the set of {\em
negative body atoms}, defined as $\nbody{r}=\set{c_1,\ldots,c_m}$.
We denote by $\body{r}=\pbody{r}\union\nbody{r}$ the set of atoms
appearing in the body of a rule $r$.
Roughly speaking, the body gives the conditions on which the head of
the rule must be satisfied.  For example, in case of a choice rule
(\ref{eq:choice-rule}), this means that any head atom 
$a_i$ can be inferred to be true if $\rg{b_1}{,}{b_n}$ hold true by
some other rules but none of the atoms $\rg{c_1}{,}{c_m}$.
Weight rules of the form (\ref{eq:weight-rule}) cover
many other kinds of rules of interest as their special cases:
\begin{eqnarray}
\label{eq:cardinality-rule}
a\IF\limit{l}{\rg{b_1}{,}{b_n},\naf\rg{c_1}{,}{\naf c_m}} \\
\label{eq:basic-rule}
a\IF\rg{b_1}{,}{b_n},\naf\rg{c_1}{,}{\naf c_m} \\
\label{eq:integrity-constraint}
\IF\rg{b_1}{,}{b_n},\naf\rg{c_1}{,}{\naf c_m} \\
\label{eq:compute-statement}
\mathsf{compute\ }\set{\rg{b_1}{,}{b_n},\rg{\naf c_1}{,}{\naf c_m}}
\end{eqnarray}
\emph{Cardinality rules} of the form (\ref{eq:cardinality-rule}) are
essentially weight rules (\ref{eq:weight-rule}) where $w=l$ and all
weights associated with literals equal to $1$.  A \emph{normal rule},
or alternatively a \emph{basic rule} (\ref{eq:basic-rule}) is a
special case of a cardinality rule (\ref{eq:cardinality-rule}) with
$l=n+m$. The intuitive meaning of an integrity constraint
(\ref{eq:integrity-constraint}) is that the conditions given in the
body are never simultaneously satisfied. The same can be stated
in terms of a basic rule
$f\IF\rg{b_1}{,}{b_n},\naf\rg{c_1}{,}{\naf c_m},\naf f$
where $f$ is a new atom dedicated to integrity constraints.
Finally, \emph{compute statements} (\ref{eq:compute-statement}) of
the \system{smodels} system effectively correspond to sets of
integrity constraints
$\rg{\IF\naf b_1}{,}{\IF\naf b_n}$ and $\rg{\IF c_1}{,}{\IF c_m}$.

Because the order of literals in (\ref{eq:weight-rule}) and
(\ref{eq:choice-rule}) is considered irrelevant, we introduce
shorthands $A=\eset{a_1}{a_h}$, $B=\eset{b_1}{b_n}$, and
$C=\eset{c_1}{c_m}$ for the sets of atoms involved in rules,
and $W_B=\eset{w_{b_1}}{w_{b_n}}$ and $W_{C}=\eset{w_{c_1}}{w_{c_m}}$
for the respective sets of weights in (\ref{eq:weight-rule}). 
Using these notations (\ref{eq:weight-rule}) and
(\ref{eq:choice-rule}) are abbreviated by $a\IF\limit{w}{B=W_B,\naf
  C=W_C}$\footnote{% 
Strictly speaking $B=W_B$ and $\naf C=W_C$ are to be
understood as sets of pairs of the form $(b,w_{b})$ and
$(\naf c,w_{c})$, respectively.
For convenience the exact matching between literals and weights is
left implicit in the shorthand.     
} and $\choice{A}\IF B,\naf C$. 

In the \system{smodels} system, the internal representation of
programs is based on rules of the forms
(\ref{eq:weight-rule})--(\ref{eq:basic-rule}) and
(\ref{eq:compute-statement}) and one may conclude that basic constraint
rules, as introduced above, provide a reasonable coverage of
\system{smodels} programs. Thus we concentrate on rules of the forms
(\ref{eq:weight-rule}) and (\ref{eq:choice-rule}) and view others as
syntactic sugar in the sequel.

\begin{definition}
\label{smodels-program}
An \system{smodels} program $P$ is a finite set of basic constraint
rules. 
\end{definition}
An \system{smodels} program consisting only of basic rules is called a 
{\em normal logic program}~(NLP), and
a basic rule with an empty body is called a {\em fact}.
Given an \system{smodels} program $P$, we write $\hb{P}$ for its
\emph{signature}, that is, the set of atoms occurring in $P$, and
$\body{P}$ and $\head{P}$ for the respective subsets of $\hb{P}$
having \emph{body occurrences} and \emph{head occurrences} in the rules
of $P$. 
Furthermore, $\choiceheads{P}\subseteq\head{P}$ denotes the set of atoms
having a head occurrence in a choice rule of $P$.

Given a program $P$, an {\em interpretation} $M$ of $P$ is a subset of 
$\hb{P}$ defining which atoms $a\in\hb{P}$ are {\em true} ($a\in M$) and
which are {\em false} ($a\not\in M$).
A weight rule (\ref{eq:weight-rule}) is satisfied in $M$ if and only
if $a\in M$ whenever the sum of weights
$$\sum_{b\in B\isect M}w_{b}+\sum_{c\in C\setminus M}w_{c}$$
is at least $w$. A choice rule $\choice{A}\IF B, \naf C$
is always satisfied in $M$. An interpretation $M\subseteq\hb{P}$ is a
(classical) model of $P$, denoted by $M\models P$, if and only if $M$
satisfies all the rules in $P$.

The generalization of the {\em Gelfond-Lifschitz
reduct}~\cite{GL88:iclp} for \system{smodels} programs is defined as
follows.  
\begin{definition}
\label{reduct}
For an \system{smodels} program $P$ and an interpretation
$M\subseteq\hb{P}$, the reduct $\GLred{P}{M}{}$ contains
\begin{enumerate}
\item
a rule $a\IF B$ if and only if there is a choice rule $\choice{A}\IF
B, \naf C$ in $P$ such that $a\in A\isect M$, and $M\isect C=\emptyset$; 
\item
a rule $a\IF\limit{w'}{B=W_{B}}$ if and only if
there is a weight rule $a\IF\limit{w}{B=W_{B},\naf C=W_{C}}$
in $P$ such that 
$w'= \max(0,w-\sum_{c\in C\setminus M}w_{c})$.
\end{enumerate}
\end{definition}
We say that an \system{smodels} program $P$ is {\em positive} if each
rule in $P$ is a weight rule restricted to the case $C=\emptyset$. 
Recalling that the basic rules are just a special case of weight
rules, we note that the reduct $\GLred{P}{M}$ is always positive.
An interpretation $M\subseteq\hb{P}$ is the {\em least model} of
a positive \system{smodels} program $P$, denoted by $\lm{P}$, if and
only if $M\models P$ and there is no $M'\models P$ such that
$M'\subset M$. 
Given the {\em least model semantics} for positive
programs~\cite{JO07:tplp}, the stable  
model semantics~\cite{GL88:iclp} straightforwardly generalizes for
\system{smodels} programs~\cite{SNS02:aij}.
\begin{definition}
\label{smodels-stable}
An interpretation $M\subseteq\hb{P}$ is a stable model of an
\system{smodels} program $P$ if and only if $M=\lm{\GLred{P}{M}{}}$.  
\end{definition}

Given an \system{smodels} program $P$ and $a,b\in\hb{P}$, we
say that $a$ {\em depends directly} on $b$, denoted by $b\leq_1 a$, if
and only if $P$ contains a rule $r$ such that $a\in\head{r}$ and 
$b\in\pbody{r}$. 
The {\em positive dependency graph} of $P$, denoted by $\dep{P}$, is
the graph $\pair{\hb{P}}{\leq_1}$. The reflexive and transitive
closure of $\leq_1$ gives rise to the dependency relation $\leq$ over
$\hb{P}$. 
A {\em strongly connected component} (SCC) $S$ of $\dep{P}$ is a
maximal set $S\subseteq\hb{P}$ such that $b\leq a$ holds for every
$a,b\in S$. 

\subsection{Splitting sets and loop formulas}
\label{section:splitting-and-loopformulas}

In this section we consider only the class of {\em normal
logic programs}. 
We formulate the {\em splitting-set theorem} \cite{LT94:iclp} in the
case of normal logic programs\footnote{
Lifschitz and Turner \shortcite{LT94:iclp} consider a more general class
of logic programs, {\em extended disjunctive logic programs}, that is,
disjunctive logic programs with two kinds of negation.},
and give an alternative definition of stable models based on the
{\em classical models} of the {\em completion} of a
program~\cite{Clark78} and its 
{\em loop formulas}~\cite{LinZ04}.   
The splitting-set theorem can be used to simplify the computation of
stable models by splitting a program into parts, and it is also a
useful tool for structuring mathematical proofs for properties of
logic programs.  

\begin{definition}
\label{def:splitting-set}
A splitting set for a normal logic program $P$ is any set
$U\subseteq \hb{P}$ 
such that for every rule $r$ in $P$ it holds that
$\hb{r}\subseteq U$ if $\head{r}\in U$. 
\end{definition}
The set of rules $r\in P$ such that $\hb{r}\subseteq U$ is the 
{\em bottom} of $P$ relative to $U$, denoted by $\bottom{P}{U}$.  
The set $\topp{P}{U}=P\setminus\bottom{P}{U}$ is the {\em top} of $P$
relative to $U$ 
which can be partially evaluated with respect to an interpretation
$X\subseteq U$. The result is a program $\rest{P}{U}{X}$
defined as 
\begin{multline*}
\{\head{r}\IF (\pbody{r}\setminus U),\naf(\nbody{r}\setminus U)
\mid r\in \topp{P}{U}, \\\pbody{r}\isect U\subseteq X\mbox{ and }
(\nbody{r}\isect U)\isect X =\emptyset\}.
\end{multline*}
A {\em solution} to a program with respect to a splitting set is a
pair consisting of a stable model~$X$ for the bottom and a stable model
$Y$ for the top partially evaluated with respect to $X$.
\begin{definition}
\label{solution}
Given a splitting set $U$ for a normal logic program $P$, a
solution to $P$ with respect to $U$ is a pair $\pair{X}{Y}$ 
such that 
\begin{enumerate}
\item[(i)] $X\subseteq U$ is a stable model of $\bottom{P}{U}$, and
\item[(ii)] $Y\subseteq\hb{P}\setminus U$ is a stable model of
$\rest{P}{U}{X}$. 
\end{enumerate}
\end{definition}
Solutions and stable models relate as follows.
\begin{theorem}[The splitting-set theorem~\cite{LT94:iclp}]
\label{thr:splitting-set} 
\ \\
Let $U$ be a splitting set for a normal logic program $P$ and
$M\subseteq\hb{P}$ an interpretation. Then $M\in\sm{P}$ if and
only if the pair $\pair{M\isect U}{M\setminus U}$ is a solution to $P$ 
with respect to $U$. 
\end{theorem}
The splitting-set theorem can also be used in an iterative manner, if
there is a {\em monotone sequence} of splitting sets $\{U_1,\ldots,
U_i,\ldots\}$, that is, $U_i\subset U_j$ if $i<j$, for program $P$.
This is called a {\em splitting sequence} and 
it induces a {\em component structure} for $P$.
The splitting-set theorem generalizes to a {\em splitting sequence
theorem}~\cite{LT94:iclp}, and given a splitting sequence, the stable
models of a program $P$ can be computed iteratively bottom-up. 

Lin and Zhao present an alternative definition of stable models for
normal logic programs based on the {\em classical models} of the
{\em completion} of a program~\cite{Clark78} and its {\em loop
  formulas}~\cite{LinZ04}.    
We will apply this definition later on in the proof of the module
theorem (Theorem \ref{moduletheorem}).

\begin{definition}[Program completion \cite{Clark78,Fages94}]
\label{clarks-completion}
The completion of a normal logic program $P$ is  
\begin{eqnarray}
\clark{P}
 =  \bigwedge_{a \in \hb{P}}
\Bigg (a \leftrightarrow
\bigvee_{\head{r}=a} 
\Bigg ( \bigwedge_{b \in \pbody{r}} b  \wedge
\bigwedge_{c \in \nbody{r}} \neg c \Bigg ) \Bigg). 
\end{eqnarray}
\end{definition}
Note that an empty body reduces to true  and in that case the
respective equivalence for an atom $a$ is logically equivalent to
$a\leftrightarrow \top$. 

\begin{definition}
\label{def:loop}
Given a normal logic program $P$, a set of atoms $L\subseteq\hb{P}$ is
a loop of $P$ if for every $a,b\in L$ there is a path of non-zero length
from $a$ to $b$ in $\dep{P}$ such that all vertices in the path are in
$L$. 
\end{definition}

\begin{definition}
\label{def:loop-formula}
Given a normal logic program $P$ and a loop $L\subseteq\hb{P}$ of $P$, 
the loop formula associated with $L$ is
$$\loops{L}{P}=
\neg\Bigg(\bigvee_{r\in\mathrm{EB}(L,P)} \Bigg (
\bigwedge_{b\in\pbody{r}}b\land\bigwedge_{c\in\nbody{r}}\neg c
\Bigg ) \Bigg )\rightarrow
\bigwedge_{a\in L}\neg a$$ 
where
$\mathrm{EB}(L,P) = \{r\in P 
\mid \head{r}\in L\mbox{ and }\pbody{r}\isect L =\emptyset\}$ is the
set of rules in $P$ which have external bodies of $L$.
\end{definition}

Now, stable models of a program and classical models of its completion
that satisfy the loop formulas relate as follows.

\begin{theorem}[\cite{LinZ04}]
\label{stable-models-using-loop-formulas}
Given a normal logic program $P$ and an interpretation
$M\subseteq\hb{P}$, $M\in\sm{P}$ if and only if
$M\models\clark{P}\union\lfs{P}$, where $\lfs{P}$ is the set
of all loop formulas associated with the loops of $P$.
\end{theorem}

\subsection{Equivalence relations for {\sc smodels} programs}
\label{sect:equivalences}

There are several notions of equivalence that have been proposed for
logic programs. 
We review a number of them in the context of \system{smodels}
programs.  

Lifschitz et al.~\shortcite{LPV01:acmtocl} address the
notions of {\em weak/ordinary equivalence} and {\em strong
equivalence}. 
\begin{definition}
\label{weak-strong-eq}
\system{smodels} programs $P$ and $Q$ are weakly equivalent,
denoted by $P\lpeq{}Q$, if and only if $\sm{P}=\sm{Q}$; and strongly
equivalent, denoted by $P\lpeq{s}Q$, if and only if $P\union
R\lpeq{}Q\union R$ for any \system{smodels} program $R$.
\end{definition}
The program $R$ in the above definition can be understood as an
arbitrary context in which the two programs being compared could
be placed. Therefore strongly equivalent logic programs are semantics
preserving substitutes of each other and relation $\lpeq{s}$ is a
{\em congruence relation} for $\union$ among \system{smodels}
programs, that is, if $P\lpeq{s}Q$, then also $P\union R\lpeq{s}Q\union R$
for all \system{smodels} programs $R$.  
Using $R=\emptyset$ as context, one sees that $P\lpeq{s}Q$ implies
$P\lpeq{}Q$. The converse does not hold in general.

A way to weaken strong equivalence is to restrict possible
contexts to sets of facts. The notion of {\em uniform equivalence} has
its roots in the database community \cite{Sagiv87}, see
\cite{EF03:iclp} for the case of the stable model semantics. 

\begin{definition}
\label{uniform-eq}
\system{smodels} programs $P$ and $Q$ are uniformly equivalent,
denoted by $P\lpeq{u}Q$, if and only if $P\union F\lpeq{}Q\union F$ 
for any set of facts $F$.
\end{definition}
Example \ref{uni-not-strong} shows that uniform equivalence is
not a congruence for union.
\begin{example}{\bf (\cite[Example 1]{EFTW04:lpnmr})}
\label{uni-not-strong}
Consider programs $P=\{a.\}$ and $Q=\{a\IF \naf b\rsep
a\IF b.\}$. It holds that $P\lpeq{u}Q$, but $P\union R\not\lpeq{}
Q\union R$ for the context $R=\{b\IF a.\}$. This implies
$P\not\lpeq{s}Q$ and $P\union R\not\lpeq{u}Q\union R$. \hfill $\bbox$  
\end{example}
There are also {\em relativized variants of strong and uniform
equivalence} \cite{Woltran04:jelia} which allow the context to  
be constrained using a set of atoms $A$.

For weak equivalence of programs $P$ and $Q$
to hold, $\sm{P}$ and $\sm{Q}$ have to be identical 
subsets of $\mathbf{2}^{\hb{P}}$ and $\mathbf{2}^{\hb{Q}}$, respectively.  
The same effect can be seen with $P\lpeq{s}Q$ and $P\lpeq{u}Q$. 
This makes these relations less useful if $\hb{P}$ and $\hb{Q}$ differ
by some (local) atoms not trivially false in all stable models.  
The {\em visible equivalence relation}~\cite{Janhunen06:jancl} 
takes the interfaces of programs into account.   
The atoms in $\hb{P}$ are partitioned into two parts, $\hbv{P}$ and
$\hbh{P}$, which determine the {\em visible} and the {\em hidden}
parts of $\hb{P}$, respectively.   
Visible atoms form an interface for interaction between programs, and
hidden atoms are local to each program and thus negligible when
visible equivalence of programs is concerned. 

\begin{definition} 
\label{equivalences} 
\system{smodels} programs $P$ and $Q$ are visibly equivalent, denoted
by $P\lpeq{v}Q$, if and only if $\hbv{P}=\hbv{Q}$ and there 
is a bijection $f\fun{\sm{P}}{\sm{Q}}$ such that for all
$M\in\sm{P}$,
$M\isect\hbv{P}=f(M)\isect\hbv{Q}$.
\end{definition}

Note that the number of stable models is also preserved under $\lpeq{v}$.
Such a strict correspondence of models is much dictated by the answer
set programming methodology: the stable models of a program usually
correspond to the solutions of the problem being solved and thus the
exact preservation of models is highly significant.
In the fully visible case, that is, for $\hbh{P}=\hbh{Q}=\emptyset$, the 
relation $\lpeq{v}$ becomes very close to $\lpeq{}$.
The only difference is the requirement that $\hb{P}=\hb{Q}$ insisted
on $\lpeq{v}$. 
This is of little importance as $\hb{P}$ can always be extended
by adding (tautological) rules of the form $a\IF a$ to $P$ without
affecting the stable models of the program. 
Since weak equivalence is not a congruence for $\union$, visible
equivalence cannot be a congruence for program union either.

The verification of weak, strong, or uniform equivalence is a
$\mathbf{coNP}$-complete decision problem for \system{smodels}
programs \cite{MT91:tplp,PTW01:epia,EF03:iclp}.
The computational complexity of deciding $\lpeq{v}$ is analyzed
in~\cite{JO07:tplp}. If the use of hidden atoms is not limited in any 
way, the problem of verifying visible equivalence becomes at least as
hard as the counting problem $\#\mathbf{SAT}$ which is
$\#\mathbf{P}$-complete \cite{Valiant79}.  
It is possible, however, to govern the computational complexity by
limiting the use of hidden atoms by the property of having {\em enough 
visible atoms}~\cite{JO07:tplp}.
Intuitively, if $P$ has enough visible atoms, the EVA property for
short, then each interpretation of $\hbv{P}$ {\em uniquely}
determines an interpretation of $\hbh{P}$. 
Consequently, the stable models of $P$ can be distinguished
on the basis of their visible parts. 
Although verifying the EVA property can be hard in general
\cite[Proposition 4.14]{JO07:tplp}, there are syntactic subclasses of
\system{smodels} programs with the EVA property.  The use of visible
atoms remains unlimited and thus the full expressiveness of
\system{smodels} programs remains at programmer's disposal.    
Also note that the EVA property can always be achieved by declaring
sufficiently many atoms visible.
For \system{smodels} programs with the EVA property, the
verification of visible equivalence is a {\bf coNP}-complete decision
problem~\cite{JO07:tplp}. 

Eiter et al.~\shortcite{ETW05:ijcai} introduce a very
general framework based on {\em equivalence frames} to capture various
kinds of equivalence relations. All the equivalence relations defined in
this section can be defined using the framework. 
Visible equivalence, however, is exceptional in the sense that it does
not fit into equivalence frames based on {\em projected answer sets}.
As a consequence, the number of answer sets may not be preserved which
is somewhat unsatisfactory because of the general nature of
answer set programming as discussed in the previous section.  
Under the EVA assumption, however, the {\em projective variant of visible
equivalence} 
defined by 
$$\set{M\isect\hbv{P}\mid M\in\sm{P}}=
\set{N\isect\hbv{Q}\mid N\in\sm{Q}}$$
coincides with visible equivalence. 

Recently Woltran presented another general framework characterizing
{\em $\tuple{\mathcal{H},\mathcal{B}}$-equivalence}~\cite{Woltran07:cent}.  
$\tuple{\mathcal{H},\mathcal{B}}$-equivalence is defined similarly to
strong equivalence, but the set of possible contexts is restricted by
limiting the head and body occurrences of atoms in a context program
$R$ by $\mathcal{H}$ and $\mathcal{B}$, respectively. 
Thus, programs $P$ and $Q$ are
$\tuple{\mathcal{H},\mathcal{B}}$-equivalent 
if and only if $P\union R\lpeq{}Q\union R$ for all
$R$ such that $\head{R}\subseteq\mathcal{H}$ and
$\body{R}\subseteq\mathcal{B}$.
Several notions of equivalence such as weak equivalence together
with (relativized) strong and (relativized) uniform equivalence can be
seen as special cases of $\tuple{\mathcal{H},\mathcal{B}}$-equivalence
by varying the sets $\mathcal{H}$ and $\mathcal{B}$.

\renewcommand{\GLred}[3]{{#1}^{#2,#3}}

%------------------------------------------------------------------------------

\section{{\sc smodels} program modules}
\label{section:modules}

We start this section by introducing the syntax and the stable
model semantics for an individual \system{smodels} program module, and
then formalize the conditions for module composition.
One of the main results is the {\em module theorem} showing that
module composition is suitably restricted so that compositionality  
of stable model semantics for \system{smodels} programs is achieved.
We also introduce an equivalence relation for modules, and propose a
general translation-based scheme for introducing syntactical
extensions for the module theorem. The scheme is then utilized in the
proof of the module theorem. 
We end this section with a brief comparison between our module
architecture and other similar proposals.

\subsection{Syntax and semantics of an {\sc smodels} program module}

We define a {\em logic program module} similarly to Gaifman and
Shapiro~\shortcite{GS89},
but consider the case of \system{smodels} programs instead of positive
normal logic programs covered in~\cite{GS89}.
An analogous module system in the context of \emph{disjunctive logic
programs} is presented in \cite{JOTW07:lpnmr}.

\begin{definition}
\label{smodels-module}
An \system{smodels} program module $\module{P}$ is a quadruple
$\tuple{R,I,O,H}$ where
\begin{enumerate}
\item $R$ is a finite set of basic constraint rules;
\item $I$, $O$, and $H$ are pairwise disjoint sets of
      input, output, and hidden atoms;
\item $\hb{R}\subseteq\hb{\module{P}}$ which is defined
      by $\hb{\module{P}}=I\union O\union H$; and 
\item $\head{R}\isect I=\emptyset$.
\end{enumerate}
\end{definition}

The atoms in $\hbv{\module{P}}=I\union O$ are considered to be
\emph{visible} and hence accessible to other modules conjoined with
$\module{P}$; either to produce input for $\module{P}$ or to utilize
the output of $\module{P}$. We use notations $\hbi{\module{P}}$ and
$\hbo{\module{P}}$ for referring to the \emph{input signature} $I$ and
the \emph{output signature} $O$, respectively. 
The \emph{hidden} atoms in
$\hbh{\module{P}}=H=\hb{\module{P}}\setminus\hbv{\module{P}}$ are used 
to formalize some auxiliary concepts of $\module{P}$ which may not be
sensible for other modules but may save space substantially.
The use of hidden atoms may yield exponential savings in space,
see~\cite[Example 4.5]{JO07:tplp}, for instance. 
The condition $\head{R}\isect I=\emptyset$ ensures that a module may
not interfere with its own input by defining input atoms of $I$ in
terms of its rules. Thus input atoms are only allowed to appear as
conditions in rule bodies.

\begin{example}
\label{ex:hc1}
Consider the Hamiltonian cycle problem for directed graphs, that is,
whether there is a cycle in the graph such that each node is visited
exactly once returning to the starting node.   

Let $n$ denote the number of nodes in the graph and let
$\mathsf{arc}(x,y)$ denote that there is a directed edge from node $x$
to node $y$ in the graph. 

Module $\module{H}^n=\tuple{R,I,O,\set{c,d}}$ selects the edges to be taken
into a cycle by insisting that each node must have exactly
one incoming and exactly one outgoing edge.  
The input signature of $\module{H}^n$ is a graph represented as a set
of edges: $I=\set{\mathsf{arc}(x,y)\mid 1\leq x,y\leq n}$. 
The output signature of $\module{H}^n$ represents which edges get
selected into a candidate for a Hamiltonian cycle: 
$O=\set{\mathsf{hc}(x,y)\mid 1\leq x,y\leq n}$.    
The set $R$ contains rules
\begin{eqnarray}
\choice{\mathsf{hc}(x,y)} &\IF& \mathsf{arc}(x,y) \label{eq:selection} \\
c &\IF& \limit{2}{\mathsf{hc}(x,1),\ldots,\mathsf{hc}(x,n)} 
\label{eq:ingoing1}\\
c &\IF& \naf\mathsf{hc}(x,1),\ldots,\naf\mathsf{hc}(x,n)\label{eq:ingoing2} \\
c &\IF&
\limit{2}{\mathsf{hc}(1,x),\ldots,\mathsf{hc}(n,x)}\label{eq:outgoing1} 
\mbox{ and}\\
c &\IF& \naf\mathsf{hc}(1,x),\ldots,\naf\mathsf{hc}(n,x)\label{eq:outgoing2} 
\end{eqnarray}
for each $1\leq x,y\leq n$; and a rule $d \IF \naf d, c$ which
enforces $c$ to be false in every stable model.
The rules in (\ref{eq:selection}) encode the selection of edges taken
in the cycle. 
The rules in (\ref{eq:ingoing1}) and (\ref{eq:ingoing2}) are used to
guarantee that each node has exactly one outgoing edge, and
the rules in (\ref{eq:outgoing1}) and (\ref{eq:outgoing2}) give the
respective condition concerning incoming edges.

We also need to check that  each node is reachable from the first
node along the edges in the cycle. For this, we introduce module
$\module{R}^n=\tuple{R',I', O', \set{e}}$. 
The input signature of $\module{R}^n$ is $I'=O$, and the output
signature is $O'=\set{\mathsf{reached}(x)\mid 1\leq x\leq n}$, where 
$\mathsf{reached}(x)$ tells that node $x$ is reachable from the first
node. 
The set $R'$ contains rules
\begin{eqnarray*}
\mathsf{reached}(y)& \IF &\mathsf{hc}(1,y)\\
\mathsf{reached}(y)& \IF &\mathsf{reached}(x), \mathsf{hc}(x,y) \\
e&\IF& \naf e, \naf\mathsf{reached}(y)
\end{eqnarray*}
for each $2\leq x\leq n$ and $1\leq y\leq n$. \hfill $\bbox$
\end{example}

To generalize the stable model semantics to cover modules as well, we
must explicate the semantical role of input atoms. To this end, we will
follow an approach%
\footnote{
There are alternative ways to handle input atoms. One possibility
is to combine a module with a set of facts (or a database) over its
input signature \cite{OJ06:ecai,Oikarinen07:lpnmr}. Yet another
approach is to interpret input atoms as \emph{fixed atoms}
in the sense of parallel circumscription~\cite{Lifschitz85:ijcai}.}
from \cite{JOTW07:lpnmr} and take input atoms into account in the
definition of the reduct adopted from \cite{JO07:tplp}. It should be
stressed that {\em all negative literals} and {\em literals involving
input atoms} get evaluated in the reduction.
Moreover, our definitions become equivalent with those proposed for
normal programs \cite{GL88:iclp} and \system{smodels} programs
\cite{JO07:tplp} if an empty input signature $I=\emptyset$ is
additionally assumed. Using the same idea, a conventional
\system{smodels} program, that is, a set of basic constraint rules $R$,
can be viewed as a module $\tuple{R,\emptyset,\hb{R},\emptyset}$
without any input atoms and all atoms visible.

\begin{definition}
\label{def:reduct}
Given a module $\module{P}=\tuple{R,I,O,H}$, the \emph{reduct} of $R$
with respect to an interpretation $M\subseteq\hb{\module{P}}$ and
input signature $I$,
denoted by $\GLred{R}{M}{I}$, contains
\begin{enumerate}
\item
a rule $a\IF (B\setminus I)$
if and only if
there is a choice rule $\choice{A}\IF B, \naf C$ in $R$ such that
$a\in A\isect M$, $B\isect I\subseteq M$, and $M\isect C=\emptyset$; and

\item
a rule $a\IF\limit{w'}{B\setminus I=W_{B\setminus I}}$ if and only if
there is a weight rule $a\IF\limit{w}{B=W_{B},\naf C=W_{C}}$
in $R$, and
$$w'=\max(0,w-\sum_{b\in B\isect I\isect M}w_b-\sum_{c\in C\setminus M}w_c).$$
\end{enumerate}
\end{definition}
As all occurrences of atoms in the input signature and
all negative occurrences of atoms are evaluated, the generalized reduct
$\GLred{R}{M}{I}$ is a 
positive program in the sense of \cite{JO07:tplp} and thus it has a
unique least model $\lm{\GLred{R}{M}{I}}\subseteq\hb{R}\setminus I$. 

\begin{definition}
\label{def:stable-model}
An interpretation $M\subseteq\hb{\module{P}}$ is a stable model of
an \system{smodels} program module $\module{P}=\tuple{R,I,O,H}$,
denoted by $M\in\sm{\module{P}}$, if and only if $M\setminus
I=\lm{\GLred{R}{M}{I}}$.  
\end{definition}

If one is interested in computing stable models of a module with
respect to a certain input interpretation, it is easier to use an
alternative definition of stable semantics for
modules~\cite{OJ06:ecai}, where   
an actual input is seen as a set of facts (or a database) to be
combined with the module. 
\begin{definition}
\label{def:instantiate-module}
Given an \system{smodels} program module $\module{P}=\tuple{R,I,O,H}$
and a set of atoms $A\subseteq I$, the instantiation of $\module{P}$
with an actual input $A$ is 
$$\module{P}(A)=\tuple{R\union \{a.\mid a\in A\}, \emptyset,
I\union O,H}.$$
\end{definition}
The module $\module{P}(A)$ is essentially an \system{smodels} program
with $I\union O$ as the set of visible atoms. Thus the stable model
semantics of \system{smodels} programs in Definition
\ref{smodels-stable} directly generalizes for an instantiated
module. 

\renewcommand{\GLred}[2]{{#1}^{#2}}

\begin{definition}
\label{def:alter-stable-model}
An interpretation $M\subseteq\hb{\module{P}}$ is a stable model of an
\system{smodels} program module $\module{P}=\tuple{R,I,O,H}$ if and
only if $M=\lm{\GLred{R}{M}\union\{a.\mid a\in M\isect I\}}.$
\end{definition}
\renewcommand{\GLred}[3]{{#1}^{#2,#3}}
It is worth emphasizing that Definitions \ref{def:stable-model} and
\ref{def:alter-stable-model} result in exactly the same semantics for 
\system{smodels} program modules. 

\begin{example}
\label{ex:hc2}
Recall module $\module{H}^n$ from Example \ref{ex:hc1}.
We consider the stable models of $\module{H}^n$ for $n=2$ to see that 
the rules in $\module{H}^n$ do not alone guarantee that each node is
reachable along the edges taken in the cycle candidate.
Consider $M=\set{\mathsf{arc}(1,1),\mathsf{arc}(2,2),\mathsf{hc}(1,1),
  \mathsf{hc}(2,2)}$. The reduct $\GLred{R}{M}{I}$ contains facts
$\mathsf{hc}(1,1)$ and $\mathsf{hc}(2,2)$; and rules
$c\IF \limit{2}{\mathsf{hc}(1,1),\mathsf{hc}(1,2)}$,
$c\IF \limit{2}{\mathsf{hc}(2,1),\mathsf{hc}(2,2)}$,
$c\IF \limit{2}{\mathsf{hc}(1,1),\mathsf{hc}(2,1)}$, and
$c\IF \limit{2}{\mathsf{hc}(1,2),\mathsf{hc}(2,2)}$; and
finally the rule $d\IF c$.
Now $M\in\sm{\module{H}^n}$ since $M=\lm{\GLred{R}{M}{I}}$.
However, $M$ does not correspond to a graph with a Hamiltonian cycle,
as node $2$ is not reachable from node $1$. \hfill $\bbox$
\end{example}

\subsection{Composing programs from  modules}
\label{section:composition}

The stable model semantics~\cite{GL88:iclp} does not lend itself
directly for program composition.  The problem is that in general,
stable models associated with modules do not determine stable models
assigned to their \emph{composition}. 

Gaifman and Shapiro~\shortcite{GS89} cover positive normal programs
under logical consequences. For their purposes, it is sufficient to
assume that whenever two modules $\module{P}_1$ and $\module{P}_2$ are
put together, their output signatures have to be disjoint and they have to
\emph{respect each other's hidden atoms}, that is,
$\hbh{\module{P}_1}\isect\hb{\module{P}_2}=\emptyset$ and
$\hbh{\module{P}_2}\isect\hb{\module{P}_1}=\emptyset$.

\begin{definition}
\label{def:program-composition}
Given \system{smodels} program modules
$\module{P}_1 = \langle R_1,I_1, O_1, H_1\rangle$ and
$\module{P}_2=\tuple{R_2,I_2,O_2,H_2}$, their composition is 
\[
\module{P}_1\oplus\module{P}_2 =
\tuple{R_1\union R_2, (I_1\setminus O_2) \union (I_2\setminus O_1),
       O_1\union O_2, H_1\union H_2}
\]
if $\hbo{\module{P}_1}\isect\hbo{\module{P}_2}=\emptyset$ and
$\module{P}_1$ and $\module{P}_2$ respect each other's hidden atoms.
\end{definition}

The following example shows that the conditions given for $\oplus$ are
not enough to guarantee compositionality in the case of stable models
and further restrictions on program composition become necessary.  
\begin{example}
\label{ex:gs-composition}
Consider normal logic program modules
$\module{P}_1=\tuple{\set{a\IF b.},\set{b},\set{a},\emptyset}$
and
$\module{P}_2=\tuple{\set{b\IF a.},\set{a},\set{b},\emptyset}$
both of which have stable models $\emptyset$ and $\set{a,b}$ by
symmetry.   
The \emph{composition} of $\module{P}_1$ and $\module{P}_2$  is 
$\module{P}_1\oplus\module{P}_2=\tuple{\set{a\IF b\rsep b\IF
a.},\emptyset,\set{a,b},\emptyset}$  and
$\sm{\module{P}_1\oplus\module{P}_2}=\{\emptyset\}$, that is,
$\set{a,b}$ is not a stable model of
$\module{P}_1\oplus\module{P}_2$.  \hfill $\bbox$
\end{example}

We define the positive dependency graph of an \system{smodels} program
module $\module{P}=\tuple{R,I,O,H}$ as $\dep{\module{P}}=\dep{R}$. 
Given that $\module{P}_1\oplus\module{P}_2$ is defined, we say that
$\module{P}_1$ and $\module{P}_2$ are \emph{mutually dependent} if and
only if
$\dep{\module{P}_1\oplus\module{P}_2}$ has an SCC $S$ 
such that
$S\isect\hbo{\module{P}_1}\neq\emptyset$ and
$S\isect\hbo{\module{P}_2}\neq\emptyset$, that is,
$S$ is \emph{shared by} $\module{P}_1$ and $\module{P}_2$.

\begin{definition}
\label{def:join}
The \emph{join} $\module{P}_1\join\module{P}_2$
of two \system{smodels} program modules $\module{P}_1$ and
$\module{P}_2$ is $\module{P}_1\oplus\module{P}_2$, provided
$\module{P}_1\oplus\module{P}_2$ is defined and $\module{P}_1$ and
$\module{P}_2$ are not mutually dependent.
\end{definition}

\begin{example}
\label{ex:join-modules}
Consider modules $\module{H}^n$ and $\module{R}^n$ from Example
\ref{ex:hc1}. 
Since $\module{H}^n$ and $\module{R}^n$ respect each other's hidden
atoms and are not mutually dependent, their join 
$\module{H}^n\join\module{R}^n=\tuple{R\union R', I, O\union O',
\set{c,d,e}}$ is defined.  \hfill $\bbox$
\end{example}

The conditions in Definition \ref{def:join} impose no
restrictions on positive dependencies {\em inside} modules or on {\em
negative} dependencies in general.  
It is straightforward to show that $\join$ has the following properties:  
\begin{enumerate}
\item[(i)]
Identity: 
$\module{P}\join\tuple{\emptyset, \emptyset, \emptyset,\emptyset} 
=\tuple{\emptyset,\emptyset, \emptyset,\emptyset}\join
\module{P}=\module{P}$ 
for all modules $\module{P}$.
\item[(ii)]
Commutativity: $\module{P}_1 \join \module{P}_2=\module{P}_2\join
\module{P}_1$ for all modules $\module{P}_1$ and $\module{P}_2$ such
that $\module{P}_1 \join \module{P}_2$ is defined. 
\item[(iii)] 
Associativity: $(\module{P}_1 \join \module{P}_2)\join \module{P}_3
=\module{P}_1\join(\module{P}_2\join \module{P}_3)$ for
all modules $\module{P}_1, \module{P}_2$ and $\module{P}_3$ such that
all pairwise joins are defined.
\end{enumerate}
The equality ``$=$'' used above denotes syntactical equality.
Also note that $\module{P}\join\module{P}$ is usually undefined,
which is a difference with respect to $\union$ for which it holds
that $P\union P=P$ for all programs $P$.
Furthermore, considering the join $\module{P}_1\join\module{P}_2$,
since each atom is defined in exactly one module, the sets of rules in
$\module{P}_1$ and $\module{P}_2$ are distinct, that is, $R_1\isect 
R_2=\emptyset$, and  
also, 
$\hb{\module{P}_1\join\module{P}_2} = \hb{\module{P}_1}\union
\hb{\module{P}_2}$,
$\hbv{\module{P}_1\join\module{P}_2} = \hbv{\module{P}_1}\union
\hbv{\module{P}_2}$, and 
$\hbh{\module{P}_1\join\module{P}_2} = \hbh{\module{P}_1}\union
\hbh{\module{P}_2}$.

Having the semantics of an individual \system{smodels} program module
now defined, we may characterize the properties of the semantics under
program composition using the notion of \emph{compatibility}. 

\begin{definition}
\label{def:compatibility}
Given \system{smodels} program modules $\module{P}_1$ and
$\module{P}_2$ such that $\module{P}_1\oplus\module{P}_2$ is defined,
we say that interpretations
$M_1\subseteq\hb{\module{P}_1}$ and $M_2\subseteq\hb{\module{P}_2}$
are compatible if and only if  
$M_1\isect\hbv{\module{P}_2}=M_2\isect\hbv{\module{P}_1}$.
\end{definition}
We use {\em natural join} $\Join$ to combine compatible
interpretations. 

\begin{definition}
\label{def:natural-join}
Given \system{smodels} program modules $\module{P}_1$ and
$\module{P}_2$ and sets of interpretations
$A_1\subseteq\mathbf{2}^{\hb{\module{P}_1}}$ and
$A_2\subseteq\mathbf{2}^{\hb{\module{P}_2}}$, the natural join 
of $A_1$ and $A_2$, denoted by $A_1\Join A_2$, is 
$$\sel{M_1\union M_2}
     {M_1\in A_1, M_2\in A_2\text{ such that }
      M_1\text{ and }M_2\text{ are compatible}}.$$
\end{definition}
The stable model semantics is compositional for $\join$, that is, 
if a program (module) consists of several submodules, its stable
models are locally stable for the respective submodules; and on the
other hand, local stability implies global stability for compatible
stable models of the submodules.  

\begin{theorem}[Module theorem \cite{Oikarinen07:lpnmr}]
\label{moduletheorem}
If $\module{P}_1$ and $\module{P}_2$ are \system{smodels} program
modules such that $\module{P}_1\join\module{P}_2$ is defined, then
$$\sm{\module{P}_1\join\module{P}_2}=\sm{\module{P}_1}\Join\sm{\module{P}_2}.$$ 
\end{theorem}
Instead of proving Theorem \ref{moduletheorem} directly from scratch
we will propose {\em a general translation-based scheme} for introducing
syntactical extensions for the module theorem. 
For this we need to define a concept of {\em modular equivalence}
first, and thus the proof of Theorem \ref{moduletheorem} is deferred
until Section \ref{proof-mod-theorem}.

It is worth noting that classical propositional theories have an
analogous property obtained by substituting $\union$ for $\join$ and
replacing stable models by classical models in Theorem
\ref{moduletheorem}, that is,
for any \system{smodels} programs $P_1$ and $P_2$,
$\cm{P_1\union P_2}=\cm{P_1}\Join\cm{P_2}$,
where $\cm{P}=\{M\subseteq\hb{P}\mid M\models P\}$.

\begin{example}
Recall modules $\module{H}^n$ and  $\module{R}^n$ in Example
\ref{ex:hc1}. In Example \ref{ex:hc2} we showed that
$M=\set{\mathsf{arc}(1,1),\mathsf{arc}(2,2),\mathsf{hc}(1,1), 
\mathsf{hc}(2,2)}$ is a stable model of $\module{H}^2$. 
Now module $\module{R}^2$ has six stable models, but none of them is 
compatible with $M$. Thus by Theorem \ref{moduletheorem}
there is no stable model $N$ for $\module{H}^2\join\module{R}^2$ such 
that $N\isect\hb{\module{H}^2}=M$. 

The join  $\module{H}^n\join\module{R}^n$ can be used to find any
graph of $n$ nodes which has a Hamiltonian cycle. For instance 
$\module{H}^2\join\module{R}^2$ has four stable models:
\begin{eqnarray*}
\set{\mathsf{arc}(1,2),\mathsf{arc}(2,1),\mathsf{hc}(1,2),\mathsf{hc}(2,1),
\mathsf{reached}(1),\mathsf{reached}(2)}\\
\set{\mathsf{arc}(1,1),\mathsf{arc}(1,2),\mathsf{arc}(2,1),\mathsf{hc}(1,2),
\mathsf{hc}(2,1),\mathsf{reached}(1),\mathsf{reached}(2)}\\
\set{\mathsf{arc}(1,2),\mathsf{arc}(2,1),\mathsf{arc}(2,2),\mathsf{hc}(1,2),
\mathsf{hc}(2,1),\mathsf{reached}(1),\mathsf{reached}(2)}\\
\set{\mathsf{arc}(1,1),\mathsf{arc}(1,2),\mathsf{arc}(2,1),\mathsf{arc}(2,2),
\mathsf{hc}(1,2),\mathsf{hc}(2,1),\mathsf{reached}(1),\mathsf{reached}(2)}.
\end{eqnarray*}
These models represent the four possible graphs of two nodes having a
Hamiltonian cycle. 
\hfill $\bbox$
\end{example}

Theorem \ref{moduletheorem} straightforwardly generalizes for
modules consisting of several submodules. Consider a collection of
\system{smodels} program modules  $\module{P}_1,\ldots,\module{P}_n$
such that the join $\module{P}_1\join\cdots\join\module{P}_n$ is
defined (recall that $\join$ is associative). 
We say that a collection of interpretations $\{M_1,\ldots,M_n\}$ for
modules $\module{P}_1,\ldots,\module{P}_n$, respectively, is {\em
compatible}, if and only if $M_i$ and $M_j$ are pairwise compatible
for all $1\leq i,j\leq n$. 
The natural join generalizes for a collection of modules as 
$$A_1\Join \cdots\Join A_n=\sel{M_1\union \cdots\union M_n}
{M_i\in A_i \text{ and }
\{M_1,\ldots,M_n\}\text{ is compatible}},$$
where $A_1\subseteq\mathbf{2}^{\hb{\module{P}_1}},\ldots,
A_n\subseteq\mathbf{2}^{\hb{\module{P}_n}}$.

\begin{corollary}
\label{general-moduletheorem}
For a collection of \system{smodels} program modules 
$\module{P}_1,\ldots,\module{P}_n$ such that the join
$\module{P}_1\join\cdots\join \module{P}_n$ is defined, it holds that
$$\sm{\module{P}_1\join\cdots\join \module{P}_n}=\sm{\module{P}_1}\Join
\cdots\Join \sm{\module{P}_n}.$$
\end{corollary}
Although Corollary \ref{general-moduletheorem} enables the computation
of stable models on a module-by-module basis, it leaves us the task
of excluding mutually incompatible combinations of stable models.
It should be noted that applying the module theorem in a naive way by
first computing stable models for each submodule and finding then the
compatible pairs afterwards, might not be preferable.

\begin{example}
\label{ex-cor-mod-thr}
Consider \system{smodels} program modules 
\begin{eqnarray*}
\module{P}_1 &=&  \tuple{\{a\IF\naf b.\},\{b\},\{a\},\emptyset},\\
\module{P}_2 &=& \tuple{\{b\IF\naf c.\},\{c\},\{b\},\emptyset}, \mbox{ and}\\
\module{P}_3 &=& \tuple{\{c\IF\naf a.\},\{a\},\{c\}, \emptyset},
\end{eqnarray*}
and their join 
$\module{P}=\module{P}_1\join\module{P}_2\join\module{P}_3
=\tuple{\{a\IF\naf b\rsep b\IF\naf c\rsep c\IF\naf
a.\},\emptyset, \{a,b,c\},\emptyset}.$
We have $\sm{\module{P}_1}=\{\{a\},\{b\}\}$,
$\sm{\module{P}_2}=\{\{b\},\{c\}\}$, and 
$\sm{\module{P}_3}=\{\{a\},\{c\}\}$. To apply 
Corollary \ref{general-moduletheorem} for finding $\sm{\module{P}}$,
a naive approach is to compute all stable models of all the
modules and try to find a compatible triple of stable models $M_1$,
$M_2$, and $M_3$ for $\module{P}_1$, $\module{P}_2$, and $\module{P}_3$,
respectively.    
\begin{itemize}
\item
Now $\{a\}\in\sm{\module{P}_1}$ and $\{c\}\in\sm{\module{P}_2}$ are
compatible, since
$\{a\}\isect\hbv{\module{P}_2}=\emptyset=\{c\}\isect\hbv{\module{P}_1}$.  
However, $\{a\}\in\sm{\module{P}_3}$ is not compatible
with $\{c\}\in\sm{\module{P}_2}$, since
$\{c\}\isect\hbv{\module{P}_3}= \{c\}\ne\emptyset=\{a\}\isect
\hbv{\module{P}_2}$. 
On the other hand, $\{c\}\in\sm{\module{P}_3}$ is not compatible with
$\{a\}\in\sm{\module{P}_1}$, since 
$\{a\}\isect\hbv{\module{P}_3}= \{a\}\ne\emptyset=\{c\}\isect
\hbv{\module{P}_1}$.
\item
Also $\{b\}\in\sm{\module{P}_1}$ and $\{b\}\in\sm{\module{P}_2}$ are
compatible, but $\{b\}\in\sm{\module{P}_1}$ is incompatible with
$\{a\}\in\sm{\module{P}_3}$. Nor is $\{b\}\in\sm{\module{P}_2}$ 
compatible with $\{c\}\in\sm{\module{P}_3}$.
\end{itemize}
Thus there are no
$M_1\in\sm{\module{P}_1}$, $M_2\in\sm{\module{P}_2}$, and
$M_3\in\sm{\module{P}_3}$ such that $\{M_1,M_2,M_3\}$ is compatible,
which is natural as $\sm{\module{P}}=\emptyset$. \hfill $\bbox$
\end{example}

It is not necessary to test all combinations of stable models
to see whether we have a compatible triple. Instead, we use the
alternative definition of stable models (Definition
\ref{def:alter-stable-model}) based on instantiating the module
with respect to an input interpretation, and apply the module theorem
similarly to the splitting-set theorem. One should notice
that the set of rules in $\module{P}$ presented in Example
\ref{ex-cor-mod-thr} has no non-trivial splitting sets, and thus the
splitting-set theorem is not applicable (in a non-trivial way) in this
case. 

\begin{example}
\label{ex-cor-mod-thr-2}
Consider \system{smodels} program modules $\module{P}_1$, 
$\module{P}_2$, and $\module{P}_3$ from Example \ref{ex-cor-mod-thr}.
Now, $\module{P}_1$ has two
stable models $M_1=\{a\}$ and $M_2=\{b\}$. 
\begin{itemize}
\item
The set $M_1\isect\hbi{\module{P}_3}=\{a\}=A_1$ can be seen as an input
interpretation for $\module{P}_3$. Module $\module{P}_3$ instantiated
with $A_1$ has one stable model: $\sm{\module{P}_3(A_1)}=\set{\set{a}}$. 
Furthermore, we can use
$A_2=\set{a}\isect\hbi{\module{P}_2}=\emptyset$  
to instantiate $\module{P}_2$: $\sm{\module{P}_2(A_2)}=\set{\set{b}}$. 
However, $\set{b}$ is not compatible with $M_1$, and thus there is no
way to find a compatible collection of stable models for the modules
starting from $M_1$.  
\item
We instantiate $\module{P}_3$ with
$M_2\isect\hbi{\module{P}_3}=\emptyset=A_3$ and get
$\sm{\module{P}_3(A_3)}=\set{\set{c}}$.
Continuing with $\set{c}\isect\hbi{\module{P}_2}=\set{c}=A_4$, we get 
$\sm{\module{P}_2(A_4)}=\set{\set{c}}$.
Again, we notice that $\set{c}$ is not compatible with $M_2$, and thus
it is not possible to find a compatible triple of stable models
starting from $M_2$ either. 
\end{itemize}
Thus we can conclude $\sm{\module{P}}=\emptyset$. \hfill $\bbox$
\end{example}

\subsection{Equivalence relations for modules}
\label{sect:mod-eq}

The notion of \emph{visible equivalence} \cite{Janhunen06:jancl} was
introduced in order to neglect hidden atoms when logic programs or
other theories of interest are compared on the basis of their models.
The compositionality property from Theorem \ref{moduletheorem} enables
us to bring the same idea to the level of program modules---giving
rise to \emph{modular equivalence} of logic programs.
Visible and modular equivalence are formulated for \system{smodels}
program modules as follows. 

\begin{definition}
\label{smodels-mod-eq}
For two \system{smodels} program modules $\module{P}$ and $\module{Q}$,
\begin{itemize}
\item
$\module{P}\lpeq{v}\module{Q}$ if and only if
$\hbv{\module{P}}=\hbv{\module{Q}}$ and there 
is a bijection $f\fun{\sm{\module{P}}}{\sm{\module{Q}}}$ such that for
all $M\in\sm{\module{P}}$,
$$M\isect\hbv{\module{P}}=f(M)\isect\hbv{\module{Q}};\mbox{ and}$$
\item
$\module{P}\lpeq{m}\module{Q}$ if and only if
$\hbi{\module{P}}=\hbi{\module{Q}}$ 
and $\module{P}\lpeq{v}\module{Q}$.
\end{itemize}
\end{definition}
We note that the condition
$\hbv{\module{P}}=\hbv{\module{Q}}$
insisted on the definition of $\lpeq{v}$, implies
$\hbo{\module{P}}=\hbo{\module{Q}}$ in the presence of
$\hbi{\module{P}}=\hbi{\module{Q}}$ as required by the
relation~$\lpeq{m}$.  Moreover, these relations coincide for
completely specified \system{smodels} programs, that is
modules~$\module{P}$ with $\hbi{\module{P}}=\emptyset$.

Modular equivalence lends itself for program substitutions in analogy
to \emph{strong equivalence} \cite{LPV01:acmtocl}, that is, the
relation $\lpeq{m}$ is a proper \emph{congruence} for the join
operator $\join$.
\begin{theorem}[Congruence]
\label{smodels-congruence}
Let $\module{P},\module{Q}$ and $\module{R}$ be \system{smodels}
program modules such that $\module{P}\join\module{R}$ and
$\module{Q}\join \module{R}$ are defined.  
If $\module{P}\lpeq{m}\module{Q}$, then
$\module{P}\join\module{R}\lpeq{m}\module{Q}\join\module{R}$. 
\end{theorem}
The proof of Theorem \ref{smodels-congruence} is given in 
\ref{proofs}. The following examples illustrate the use of modular
equivalence in practice. 

\begin{example}
Recall programs $P=\{a.\}$ and $Q=\{a\IF\naf b\rsep a\IF b.\}$ from 
Example \ref{uni-not-strong}. We can define modules based on them: 
$\module{P}=\tuple{P, \{b\},\{a\}, \emptyset}$ and
$\module{Q}=\tuple{Q, \{b\},\{a\},\emptyset}$. Now 
it is impossible to define a module $\module{R}$ based on $R=\{b\IF
a.\}$ in a way that $\module{Q}\join\module{R}$ would be defined.
Moreover, it holds that $\module{P}\lpeq{m}\module{Q}$.  
\hfill $\bbox$
\end{example}

\begin{example}
Module $\module{HR}^n=\tuple{R'', I'', O'',\set{f}}$ is based on an
alternative encoding for Hamiltonian cycle problem given
in~\cite{SNS02:aij}. 
In contrast to the encoding described in Example \ref{ex:hc1}, this
encoding does not allow us to separate the selection of the edges to
the cycle and the checking of reached vertices into separate
modules as their definitions are mutually dependent.
The input signature of $\module{HR}^n$ is the same as for
$\module{H}^n$, that is,  
$I''=I=\set{\mathsf{arc}(x,y)\mid 1\leq x,y\leq n}$.
The output signature of  $\module{HR}^n$ is the output signature of
$\module{H}^n\join\module{R}^n$, that is, 
$$O''=O\union O'=\set{\mathsf{hc}(x,y)\mid 1\leq x,y\leq n}\union
\set{\mathsf{reached}(x)\mid 1\leq x\leq n}.$$
The set $R''$ contains rules
\begin{eqnarray}
\choice{\mathsf{hc}(1,x)}&\IF& \mathsf{arc}(1,x) \nonumber \\
\choice{\mathsf{hc}(x,y)}&\IF &\mathsf{reached}(x),\mathsf{arc}(x,y)
\nonumber\\  
\mathsf{reached}(y) &\IF& \mathsf{hc}(x,y)\nonumber\\
f&\IF& \naf f, \naf\mathsf{reached}(x)  \nonumber\\
f&\IF& \naf f, \mathsf{hc}(x,y), \mathsf{hc}(x,z) \mbox{ and}
\label{y-ne-z}\\
f&\IF& \naf f, \mathsf{hc}(x,y), \mathsf{hc}(z,y)
\label{x-ne-z}
\end{eqnarray}
for each $1\leq x,y,z\leq n$ such that $y\ne z$ in (\ref{y-ne-z}) and
$x\ne z$ in (\ref{x-ne-z}).

Now, one may notice that $\module{HR}^n$ and
$\module{H}^n\join\module{R}^n$ have the same input/output interface, and
$\sm{\module{HR}^n}=\sm{\module{H}^n\join\module{R}^n}$
which implies $\module{HR}^n\lpeq{m}\module{H}^n\join\module{R}^n$.
\hfill $\bbox$
\end{example}

As regards the relationship between modular equivalence and
previously proposed notions of equivalence, we note the following. 
First, if one considers the {\em fully visible case}, that is,
the restriction $\hbh{\module{P}}=\hbh{\module{Q}}=\emptyset$,
modular equivalence can be seen as a special case of $A$-uniform
equivalence for $A=I$. Recall, however, the restriction that input
atoms may not appear in the heads of the rules as imposed by module
structure.     
With a further restriction $\hbi{\module{P}}=\hbi{\module{Q}}
=\emptyset$, modular equivalence  basically coincides with weak
equivalence because $\hb{\module{P}}=\hb{\module{Q}}$ can always be
satisfied by extending the interface of the module.
Setting $\hbi{\module{P}}=\hb{\module{P}}$ would in principle give us
uniform equivalence, but the additional condition $\head{R}\isect
I=\emptyset$ leaves room for the empty module only.  

In the general case with hidden atoms, the problem of verifying
$\lpeq{m}$ for \system{smodels} program modules can be reduced to
verifying $\lpeq{v}$ for \system{smodels} programs. 
This is achieved by introducing a special module $\module{G}_I$ 
containing a single choice rule, which acts as a context generator in
analogy to \cite{Woltran04:jelia}.  
We say that two modules $\module{P}$ and $\module{Q}$ are {\em
compatible} if they have the same input/output interface,
that is, $\hbi{\module{P}}=\hbi{\module{Q}}$ and 
$\hbo{\module{P}}=\hbo{\module{Q}}$.

\begin{lemma}
\label{reduce-modular-to-visible}
Consider compatible \system{smodels} program modules
$\module{P}$ and $\module{Q}$. Now $\module{P}\lpeq{m}\module{Q}$ if
and only if 
$\module{P}\join\module{G}_I\lpeq{v}\module{Q}\join\module{G}_I$ where
$I=\hbi{\module{P}}=\hbi{\module{Q}}$ and
$\module{G}_I=\tuple{\{\choice{I}\IF\},\emptyset, I, \emptyset}$
generates all possible input interpretations for $\module{P}$ and
$\module{Q}$. 
\end{lemma}

\begin{proof}
Notice that $\module{P}\join\module{G}_I$ and $\module{Q}\join
\module{G}_I$ are \system{smodels} program modules with empty input
signatures, and thus they can also be viewed as \system{smodels}
programs. 
($\implies$) 
Assume $\module{P}\lpeq{m}\module{Q}$. Since
$\module{P}\join\module{G}_I$ and $\module{Q}\join\module{G}_I$ are
defined,
$\module{P}\join\module{G}_I\lpeq{m}\module{Q}\join\module{G}_I$  
by Theorem \ref{smodels-congruence}. This implies
$\module{P}\join\module{G}_I\lpeq{v}\module{Q}\join\module{G}_I$. 
($\impliedby$)
Assume
$\module{P}\join\module{G}_I\lpeq{v}\module{Q}\join\module{G}_I$, that
is, 
$\hbv{\module{P}}=\hbv{\module{Q}}$ and there is a bijection $f:
\sm{\module{P}\join\module{G}_I}\rightarrow\sm{\module{Q}\join\module{G}_I}$
such that for each $M\in\sm{\module{P}\join\module{G}_I}$,
$M\isect\hbv{\module{P}}=f(M)\isect\hbv{\module{Q}}$. 
By Theorem~\ref{moduletheorem},
$\sm{\module{P}\join\module{G}_I}=\sm{\module{P}}\Join\sm{\module{G}_I}$
and
$\sm{\module{Q}\join\module{G}_I}=\sm{\module{Q}}\Join\sm{\module{G}_I}$.
Now,
$\sm{\module{G}_I}=\mathbf{2}^{I}$, and thus
$\sm{\module{P}\join\module{G}_I}= \sm{\module{P}}$ and
$\sm{\module{Q}\join\module{G}_I}=\sm{\module{Q}}$. This implies 
$\module{P}\lpeq{v}\module{Q}$, and
furthermore $\module{P}\lpeq{m}\module{Q}$ since $\module{P}$ and
$\module{Q}$ are compatible \system{smodels} program modules.
\end{proof}

Due to the close relationship of $\lpeq{v}$ and $\lpeq{m}$, the
respective verification problems have the same computational
complexity. As already observed in \cite{JO07:tplp}, the verification
of $\module{P}\lpeq{v}\module{Q}$ involves a \emph{counting problem}
in general and, in particular, if
$\hbv{\module{P}}=\hbv{\module{Q}}=\emptyset$.
In this special setting $\module{P}\lpeq{v}\module{Q}$ holds if and
only if $|\sm{\module{P}}|=|\sm{\module{Q}}|$, that is, the numbers of
stable models for $\module{P}$ and $\module{Q}$ coincide.
A reduction of computational time complexity can be achieved
for modules that have \emph{enough visible atoms}, that is, the EVA
property.
Basically, we say that module $\module{P}=\tuple{R,I,O,H}$ has enough
visible atoms, if and only if $R$ has enough visible atoms with
respect to $\hbv{P}=I\union O$.
However, the property of having enough visible atoms can be elegantly
stated using modules.  
We define the \emph{hidden part} of a module
$\module{P}=\tuple{R,I,O,H}$ as
$\hid{\module{P}}=\tuple{\hid{R},I\union O,H,\emptyset}$
where $\hid{R}$ contains all rules of $R$ involving atoms of $H$ in 
their heads. For a choice rule $\choice{A}\IF B,\naf C\in R$, we
take the projection $\choice{A\isect H}\IF B,\naf C$ in $\hid{R}$. 

\begin{definition}[The EVA property \cite{JO07:tplp}]
\label{EVA-modules}
\ \\
An \system{smodels} program module
$\module{P}=\tuple{R,I,O,H}$
has enough visible atoms if and only if the hidden part
$\hid{\module{P}}=\tuple{\hid{R},I\union O,H,\emptyset}$
has a unique stable model $M$ for each interpretation
$N\subseteq\hbv{\module{P}}=I\union O$
such that $M\isect(I\union O)=N$.
\end{definition}

Verifying the EVA property is $\mathbf{coNP}$-hard and
in $\Pi^\mathbf{P}_2$ for \system{smodels} programs~\cite[Proposition
4.14]{JO07:tplp}, and thus for \system{smodels} program modules, too.    
It is always possible to enforce the EVA property by
uncovering sufficiently many hidden atoms: a module $\module{P}$ for
which $\hbh{\module{P}}=\emptyset$ has clearly enough visible atoms
because $\hid{\module{P}}$ has no rules.
It is also important to realize that choice rules involving hidden
atoms in their heads most likely break up the EVA property---unless
additional constraints are introduced to exclude multiple 
models created by choices. 

Based on the observations we can conclude that verifying the
modular equivalence of modules with the EVA property is a
$\mathbf{coNP}$-complete decision problem. 
Motivated by the complexity result and by previous proposals for
translating various equivalence verification problems into the problem
of computing stable models (see
\cite{JO02:jelia,Turner03:tplp,Woltran04:jelia} for instance), we
recently introduced a translation-based method for verifying modular 
equivalence~\cite{OJ08:jlc}.   
In the following theorem, $\eqt(\cdot,\cdot)$ is the linear translation
function mapping two \system{smodels} program modules into one
\system{smodels} program module presented
in~\cite[Definition~10]{OJ08:jlc}.  
\begin{theorem} {\bf (\cite[Theorem 4]{OJ08:jlc})}
\label{eq-test-with-context}
Let $\module{P}$ and $\module{Q}$ be compatible \system{smodels}  
program modules with the EVA property, and $\module{C}$ any
\system{smodels} program module such that
$\module{P}\join\module{C}$ and $\module{Q}\join\module{C}$ are
defined.    
Then $\module{P}\join\module{C}\lpeq{m}\module{Q}\join\module{C}$ if
and only if
$\sm{\eqt(\module{P},\module{Q})\join\module{C}}=
\sm{\eqt(\module{Q},\module{P})\join\module{C}}=\emptyset$.
\end{theorem}

\subsection{Proving the module theorem using a general
  translation-based extension scheme}
\label{proof-mod-theorem}

Let us now proceed to the proof of the module theorem. We describe the
overall strategy in this section whereas detailed proofs for the theorems
are provided in \ref{proofs}. 
Instead of proving Theorem \ref{moduletheorem} from scratch, we first 
show that the theorem holds for normal logic program modules, and then
present a general scheme that enables us to derive extensions of the
module theorem syntactically in terms of translations.  

We start by stating the module theorem for normal logic program modules.
\begin{theorem}[\cite{OJ06:ecai}]
\label{theorem:nlp-moduletheorem}
If $\module{P}_1$ and $\module{P}_2$ are normal logic program modules
such that $\module{P}_1\join\module{P}_2$ is defined, then 
$$\sm{\module{P}_1\join\module{P}_2}=\sm{\module{P}_1}\Join\sm{
\module{P}_2}.$$    
\end{theorem}
Proof for Theorem \ref{theorem:nlp-moduletheorem} is given in
\ref{proofs}.  

Next definition states the conditions which we require a translation
function to have in order to achieve syntactical extensions to the
module theorem.
Intuitively, the conditions serve the following purposes:  
first, the translation has to be {\em strongly faithful}, that is, it
preserves the roles of all atoms in the original module;
second, it is {\em $\join$-preserving}, that is, possible compositions of
modules are not limited by the translation; and
third, the translation is {\em modular}.

For convenience, we define an operator
$\reveal{\module{P},A}=\tuple{R, I, O\union A, H\setminus A}$ for any  
program module $\module{P}=\tuple{R, I, O, H}$ and for any set of
atoms $A\subseteq H$. 
The revealing operator is used to make a set of hidden atoms of a module 
visible to other modules.  

\begin{definition}
\label{def:conditions-for-translation}
Let $\mathcal{C}_1$ and $\mathcal{C}_2$ be two classes of logic
program modules such that $\mathcal{C}_2\subseteq\mathcal{C}_1$.
A translation function $\trop{}: \mathcal{C}_1\rightarrow
\mathcal{C}_2$ is {\em strongly faithful, modular and
$\join$-preserving}, if the following hold for any program modules 
$\module{P},\module{Q}\in\mathcal{C}_1$: 
\begin{enumerate}
\item 
$\reveal{\module{P},\hbh{\module{P}}}\lpeq{m}\reveal{\tr{}{\module{P}},
\hbh{\module{P}}}$;
\item 
if $\module{P}\join\module{Q}$ is defined, then   
$\tr{}{\module{P}}\join\tr{}{\module{Q}}$ is defined; and
\item 
$\tr{}{\module{P}}\join\tr{}{\module{Q}}=%\lpeq{m}
\tr{}{\module{P}\join\module{Q}}$.
\end{enumerate}
\end{definition}

Notice that the condition for strong faithfulness requires
$\hbi{\tr{}{\module{P}}}=\hbi{\module{P}}$,
$\hbo{\tr{}{\module{P}}}\union\hbh{\module{P}}=\hbo{\module{P}}
\union\hbh{\module{P}}$, and
$\hbh{\module{P}}\subseteq\hbh{\tr{}{\module{P}}}$ 
to hold. 
Moreover, strong faithfulness implies {\em faithfulness}, that is, 
$\module{P}\lpeq{m}\tr{}{\module{P}}$.

\begin{theorem}
\label{theorem:modulethr-translation}
Let $\mathcal{C}_1$ and $\mathcal{C}_2$ be two classes of logic
program modules such that $\mathcal{C}_2\subseteq\mathcal{C}_1$ and
there is a translation function $\trop{}\!: \mathcal{C}_1\rightarrow
\mathcal{C}_2$ that is strongly faithful, 
$\join$-preserving, and modular as given in Definition
\ref{def:conditions-for-translation}. 
If the module theorem holds for modules in $\mathcal{C}_2$, then
it holds for modules in $\mathcal{C}_1$. 
\end{theorem}
The proof of Theorem~\ref{theorem:modulethr-translation} is provided in
\ref{proofs}. 

As regards the translation from \system{smodels} program modules to
NLP modules, it suffices, for example, to take a natural translation
similarly to~\cite{SNS02:aij}.  
Note that the translation presented in Definition \ref{smodels2basic} is
in the worst case exponential with respect to the number of rules in the
original module. For a more compact translation, see 
\cite{FL05:tplp}, for example. 

\begin{definition}
\label{smodels2basic}
Given an \system{smodels} program module $\module{P}=\tuple{R,I,O,H}$, its
translation into a normal logic program module is 
$\tr{NLP}{\module{P}}=\tuple{R', I, O, H\union H'}$, where
$R'$ contains the following rules: 
\begin{itemize}
\item
for each choice rule $\{A\}\IF B, \naf C\in R$  the set of rules 
$$\{a\IF B, \naf C,\naf \overline{a}\rsep\;
\overline{a}\IF \naf a\mid a\in A\};$$
\item
for each weight rule $a\IF\limit{w}{B=W_{B},\naf C=W_{C}}\in R$ the
set of rules  
$$\{a\IF B', \naf C'\mid B'\subseteq B, C'\subseteq C
\mbox{ and }w\leq \sum_{b\in B'}w_b +  
\sum_{c\in C'}w_c\},$$
\end{itemize}
where each $\overline{a}$ is a new atom not appearing in
$\hb{\module{P}}$ and $H'=\{\overline{a}\mid a\in\choiceheads{R}\}$.
\end{definition}

\begin{theorem}
\label{theorem:translation-smodels2nlp}
The translation $\trop{NLP}$ from \system{smodels} program modules to
normal logic program modules given in Definition \ref{smodels2basic} 
is strongly faithful, $\join$-preserving, and modular.
\end{theorem}
The proof of Theorem \ref{theorem:translation-smodels2nlp} is given in
\ref{proofs}.

The module theorem now directly follows from Theorems
\ref{theorem:nlp-moduletheorem},
\ref{theorem:modulethr-translation}, and 
\ref{theorem:translation-smodels2nlp}. 

\begin{proof}[Proof of Theorem \ref{moduletheorem}]
By Theorem \ref{theorem:nlp-moduletheorem} we know that the module
theorem holds for normal logic program modules.
Theorem \ref{theorem:modulethr-translation} shows that 
Definition \ref{def:conditions-for-translation} gives the conditions
under which Theorem \ref{theorem:nlp-moduletheorem} can be directly
generalized for a larger class of logic program modules.
By Theorem \ref{theorem:translation-smodels2nlp} we know that the 
translation $\trop{NLP}$ from \system{smodels} program modules to NLP
modules introduced in Definition \ref{smodels2basic} satisfies the
conditions given in Definition \ref{def:conditions-for-translation},
and therefore \system{smodels} program modules are covered by
the module theorem.
\end{proof}

\subsection{Comparison with earlier approaches}
\label{sect:compare-module-system}

Our module system resembles the module system proposed in~\cite{GS89}.
However, to make our system compatible with the stable model semantics
we need to introduce a further restriction of mutual dependence, that
is, we need to deny positive recursion between modules. 
Also other propositions involve similar conditions for module
composition.  
For example, Brogi et al.~\shortcite{BMPT94} employ visibility
conditions that correspond to respecting hidden atoms. However, their
approach covers only positive programs under the least model semantics. 
Maher~\shortcite{Maher93} forbids all recursion between modules and
considers Przymusinski's {\em perfect models} \cite{Przymusinski88}
rather than stable models. 
Etalle and Gabbrielli \shortcite{EG96:tcs} restrict the composition of
{\em constraint logic program}~\cite{JM94} modules with a
condition that is close to ours:
$\hb{P}\isect\hb{Q}\subseteq\hbv{P}\isect\hbv{Q}$ 
but no distinction between input and output is made, for example,
$\hbo{P}\isect\hbo{Q}\neq\emptyset$ is allowed according to their
definitions. 

Approaches to modularity within ASP typically do not allow any
recursion (negative or positive) between modules
\cite{EGM97:acm,TBA05:asp,LT94:iclp,GG99}.  
Theorem \ref{moduletheorem}, the module theorem, is strictly stronger
than the splitting-set theorem~\cite{LT94:iclp} for normal logic
programs, and the general case allows us to generalize the
splitting-set theorem for \system{smodels} programs. 
Consider first the case of normal logic programs. A {\em splitting}
of a program can be used as a basis for a module structure. 
If $U$ is a splitting set for a normal logic program $P$, then we can
define   
$$P=\module{B}\join\module{T}=\tuple{\bottom{P}{U},\emptyset,
U,\emptyset}\join\tuple{\topp{P}{U}, U, \hb{P}\setminus
U,\emptyset}.$$  
It follows directly from Theorems \ref{thr:splitting-set} and
\ref{moduletheorem} that $M_1\in\sm{\module{B}}$ and
$M_2\in\sm{\module{T}}$ are compatible if and only if $\langle
M_1,M_2\setminus U\rangle$ is a solution for $P$ with respect to $U$.

\begin{example}
\label{modulethr-vs-splitting}
Consider a normal logic program $P=\{a\IF\naf b\rsep b\IF \naf a\rsep
c\IF a.\}.$  
The set $U=\{a,b\}$ is a splitting set for $P$, and therefore
the splitting set-theorem (Theorem \ref{thr:splitting-set}) can be applied:
$\bottom{P}{U}=\{a\IF\naf 
b\rsep b\IF \naf a.\}$ and $\topp{P}{U}=\{c\IF a.\}$. 
Now $M_1=\{a\}$ and $M_2=\{b\}$ are the stable models of $\bottom{P}{U}$,
and we can evaluate the top with respect to $M_1$ and $M_2$, resulting
in solutions 
$\pair{M_1}{\{c\}}$ and $\pair{M_2}{\emptyset}$, respectively.
On the other hand, $P$ can be seen as join of modules
$\module{P}_1=\tuple{\bottom{P}{U}, \emptyset, U,\emptyset}$ and 
$\module{P}_2=\tuple{\topp{P}{U}, U,\{c\},\emptyset}$. 
Now, we have $\sm{\module{P}_1}=\{M_1,M_2\}$
and $\sm{\module{P}_2}=\{\emptyset,\{b\},\{a,c\},\{a,b,c\}\}$. 
Out of eight possible pairs only
$\pair{M_1}{\{a,c\}}$ and $\pair{M_2}{\{b\}}$ are compatible.  

However, it is possible to apply Theorem \ref{moduletheorem} similarly
to the splitting-set theorem, that is,
we only need to compute the stable models of $\module{P}_2$ 
compatible with the stable models of $\module{P}_1$. 
Notice that when the splitting-set theorem is applicable, the
stable models of $\module{P}_1$ fully define the possible input
interpretations for $\module{P}_2$.
This leaves us with stable models $\{a,c\}$ and  $\{b\}$ for the
composition. \hfill $\bbox$ 
\end{example}

On the other hand, consider the module
$\module{P}_1=\tuple{\bottom{P}{U}, \emptyset, U,\emptyset}$ in the
above example.
There are no non-trivial splitting sets for the bottom program
$\bottom{P}{U}=\{a\IF \naf b.\;b\IF \naf a.\}$.  
However, $\module{P}_1$ can be viewed as the join of two NLP modules
$\module{Q}_1 =  \tuple{\{a\IF \naf b.\}, \{b\}, \{a\},\emptyset}$,
and
$\module{Q}_2 = \tuple{ \{b\IF \naf a.\}, \{a\}, \{b\}, \emptyset}$
to which the module theorem is applicable.

In the general case of \system{smodels} program modules we can use the 
module theorem to generalize the splitting-set theorem for
\system{smodels} programs. Then the bottom module acts as an input 
generator for the top module, and one can simply find the stable
models for the top module instantiated with the stable models of the
bottom module.  
The latter strategy used in Example \ref{modulethr-vs-splitting} works
even if there is negative recursion between the modules, as already
shown in Example \ref{ex-cor-mod-thr-2}.

The module theorem strengthens an earlier version given in
\cite{Janhunen06:jancl} to cover programs that involve positive body
literals, too.     
The independent sets proposed by Faber et al.~\shortcite{FGL05:icdt} push
negative recursion inside modules which is unnecessary in view of our
results. Their version of the module theorem is also weaker
than Theorem \ref{moduletheorem}.

The approach to modularity based on
lp-functions~\cite{GG99,Baral:knowledge} has features similar to our
approach. 
The components presented by lp-functions have an input/output
interface and a {\em domain} reflecting the possible input interpretations.
The functional specification requires an lp-function to have a
{\em consistent answer set} for any interpretation in its domain.
This is something that is not required in our module system. 
Lp-functions are flexible in the sense that there are several
operators for refining them. However, the composition operator for
lp-functions allows only incremental compositions, which again
basically reflects the splitting-set theorem. 

%------------------------------------------------------------------------------

\section{More on program (de)composition}
\label{section:decomposition-and-semantical-join}

So far we have established a module architecture for
the class \system{smodels} programs, in which modules interact through
an input/output interface and the stable model semantics is
fully compatible with the architecture.
In this section we investigate further the ways to understand the
internal structure of logic programs by seeing  them as compositions of
logic program modules.
First, we use the conditions for module composition to introduce a
method for decomposing an \system{smodels} program into modules.
A more detailed knowledge of the internal structure of a program (or a
module) might reveal ways to improve search for stable models. Another
application can be found in modularization of the translation-based
equivalence verification method in~\cite{OJ08:jlc}.
Second, we consider possibilities of relaxing the conditions for
module composition, that is, whether it is possible to allow positive
recursion between modules in certain cases.     

\subsection{Finding a program decomposition}
\label{section:decomposition}

Recall that any \system{smodels} program $P$ can be viewed as a module 
$\tuple{P,\emptyset, \hb{P}, \emptyset}$, and thus we consider here a
more general case of finding a module decomposition for an
arbitrary \system{smodels} program module $\module{P}=\tuple{R,I,O,H}$.
The first step is to exploit the strongly connected components
$D_1,\ldots,D_n$ of $\dep{\module{P}}$   
and define submodules~$\module{P}_i$ by grouping the rules so that for
each $D_i$ all the rules $r\in R$ such that $\head{r}\subseteq D_i$
are put into one submodule. 

Now, the question is whether $\module{P}_i$'s defined this way would
form a valid decomposition of $\module{P}$ into submodules. 
First notice that input atoms form a special case because
$\head{R}\isect I=\emptyset$. Each $a\in I$ ends up in its own
strongly connected component and there are no rules to include into a
submodule corresponding to strongly connected component $\{a\}$. 
Thus it is actually unnecessary to include a submodule based on such a
component. 
Obviously, each weight rule in $R$ goes into exactly one of the
submodules.    
One should notice that for a choice rule $r\in R$ it can happen that
$\head{r}\isect D_i\ne\emptyset$ and $\head{r}\isect D_j\ne\emptyset$
for $i\ne j$. 
This is not a problem, since it is always possible to {\em split a
choice rule} by projecting the head, that is, by replacing a choice
rule of the form $\choice{A}\IF B,\naf C$ with choice rules
$\choice{A\isect D_i}\IF B,\naf C$ for each SCC $D_i$ such that
$A\isect D_i\ne\emptyset$.\footnote{
Note that in the case of disjunctive logic programs, splitting a rule
into two modules is more involved, see~\cite{JOTW07:lpnmr} for a
discussion on a general shifting principle.} 

Based on the discussion above, we define the {\em set of rules
defining a set of atoms} for an \system{smodels} program module.

\begin{definition}
\label{rule-set-defining}
Given an \system{smodels} program module $\module{P}=\tuple{R,I,O,H}$
and a set of atoms $D\subseteq\hb{\module{P}}\setminus I$, the set of
rules defining $D$, denoted by $R[D]$, contains the following rules:
\begin{itemize}
\item
a choice rule $\choice{A\isect D}\IF B,\naf C$ if and only if
there is a choice rule~$\choice{A}\IF B,\naf C$ in $R$
such that $A\isect D\ne\emptyset$; and
\item
a weight rule $a\IF\limit{w}{B=W_B,\naf C=W_C}$ if and only if 
there is a weight rule~$a\IF\limit{w}{B=W_B,\naf C=W_C}$ in $R$ such
that $a\in D$.
\end{itemize}
\end{definition}

We continue by defining a submodule of $\module{P}=\tuple{R,I,O,H}$ 
induced by a set of atoms $D\subseteq\hb{\module{P}}\setminus I$. We
use Definition \ref{rule-set-defining} for the set of rules, and
choose $D\isect O$ to be the output signature and the rest of the
visible atoms appearing in $R[D]$ to be the input signature.

\begin{definition}
\label{induced-module}
Given an \system{smodels} program module $\module{P}=\tuple{R,I,O,H}$
and a set of atoms $D\subseteq\hb{\module{P}}\setminus I$, a submodule
induced by $D$ is
$$\module{P}[D]=(R[D], 
(\hb{R[D]}\setminus D)\isect(I\union O), 
D\isect O,
D\isect H).$$
\end{definition}
Let $D_{1},\ldots D_{m}$ be the strongly connected components of 
$\dep{\module{P}}$ such that $D_{i}\isect I=\emptyset$.
Now we can define 
$\module{P}_i=\module{P}[D_i]$ for each $1\leq i\leq m$.
Since the strongly connected components of $\dep{\module{P}}$ are used as
a basis, it is guaranteed that there is no positive recursion between
any of the submodules $\module{P}_i$. 
Also, it is clear that the output signatures of the submodules are
pairwise disjoint. 
Unfortunately this construction does not yet guarantee that
hidden atoms stay local, and therefore the composition
$\module{P}_1\oplus\cdots\oplus\module{P}_m$
might not be defined because certain $\module{P}_i$'s
might not respect each others hidden atoms.  

A solution is to combine $D_{i}$'s in a way that modules will be
closed with respect to dependencies caused by the hidden atoms, that is, 
if a hidden atom $h$ belongs to a component $D_{i}$, then also all the 
atoms in the heads of rules in which $h$ or $\naf h$ appears, have to
belong to $D_{i}$, too.  
This can be achieved by finding the strongly connected components,
denoted by $E_1,\ldots,E_k$, for 
$\mathrm{Dep}^\mathrm{h}(\module{P},\{D_{1},\ldots,D_{m}\})$,  
where
$\mathrm{Dep}^\mathrm{h}(\module{P},\{D_{1},\ldots,D_{m}\})$ has
$\{D_{1},\ldots,D_{m}\}$ as the set of vertices, and
\begin{multline*}
\{\pair{D_{i}}{D_{j}}, \pair{D_{j}}{D_{i}}\mid 
a\in D_{i}, b\in D_{j}, r\in R, \\
b\in\head{r}\mbox{ and }a\in\body{r}\isect\hbh{\module{P}}\}   
\end{multline*}
as the set of edges. 
Now, we take the sets $$F_i=\bigcup_{D\in E_i}D$$ for
$1\leq i\leq k$ and use them to induce a module structure for
$\module{P}$ by defining $\module{P}_i=\module{P}[F_i]$ for
$1\leq i\leq k$.

As there may be atoms in $\hb{\module{P}}$ not appearing in
the rules of $\module{P}$, that is, $\hb{\module{P}}=\hb{R}$
does not necessarily hold for $\module{P}=\tuple{R,I,O,H}$, it is
possible that  
$$\hb{\module{P}}\setminus(\hb{\module{P}_1}\union\cdots\union
\hb{\module{P}_k})\ne\emptyset.$$   
To keep track of such atoms in $I\setminus\hb{R}$ we need an
additional module defined as 
$$\module{P}_0=\tuple{\emptyset,
  I\setminus\hb{R},\emptyset,\emptyset}.$$   
There is no need for a similar treatment for atoms in $(O\union
H)\setminus\hb{R}$ as each atom in $O\union H$ belongs to 
some $\hb{\module{P}_i}$ by definition. 

Theorem \ref{decomposition} shows that we have a valid decomposition
of $\module{P}$ into submodules.
\begin{theorem}
\label{decomposition}
Consider an \system{smodels} program module $\module{P}$, and
let $D_{1},\ldots D_{m}$ be the SCCs of 
$\dep{\module{P}}$ such that $D_{i}\isect I=\emptyset$, and
$E_1,\ldots,E_k$ the strongly connected components of
$\mathrm{Dep}^\mathrm{h}(\module{P},\{D_{1},\ldots,D_{m}\})$.
Define 
$\module{P}_0=\tuple{\emptyset, I\setminus\hb{R},
\emptyset,\emptyset}$, and
$\module{P}_i=\module{P}[F_i]$ for
$F_i=\bigcup_{D\in E_i}D$ and $1\leq i\leq k$.
Then the join of the submodules $\module{P}_i$ for $0\leq i\leq k$ is 
defined and 
$\module{P}\lpeq{m}\module{P}_0\join\cdots\join\module{P}_k$.
\end{theorem}

\begin{proof}
Based on the construction of $F_i$'s and the discussion in
this section it is clear that
$\module{P'}=\module{P}_0\join\cdots\join\module{P}_k$ is defined.
It is easy to verify that the sets of input, output, and hidden atoms
of modules $\module{P'}$ and $\module{P}$ are exactly the same.
The only difference between the sets of rules in $\module{P}$ and
$\module{P'}$ is that some choice rules in $\module{P}$ may have been
split into several rules in $\module{P'}$. 
This is a syntactical change not affecting the stable models of
the modules, that is, $\sm{\module{P}}=\sm{\module{P}'}$. 
Notice also that $\dep{\module{P}}=\dep{\module{P}'}$. Thus
it holds that $\module{P'}\lpeq{m}\module{P}$.
\end{proof}

\subsection{Semantical conditions for module composition}
\label{section:sem-mod-eq}

Even though Example \ref{ex:gs-composition} shows that
conditions for $\oplus$ are not enough to guarantee that the module
theorem holds, there are cases where $\module{P}\join\module{Q}$ is
not defined and still it holds that
$\sm{\module{P}\oplus\module{Q}}=\sm{\module{P}}\Join\sm{\module{Q}}$. 

\begin{example}
\label{ex3}
Consider modules $\module{P}=\tuple{\{a\IF b \rsep
a\IF\naf c.\}, \{b\}, \{a,c\},\emptyset}$ and  
$\module{Q}=\tuple{\{b\IF a.\}, \{a\}, \{b\},\emptyset}$. 
Now, the composition
$$\module{P}\oplus\module{Q}=\tuple{\{a\IF b \rsep a\IF\naf c\rsep b\IF
a.\},\emptyset,\{a,b,c\}, \emptyset}$$ 
is defined as the output sets
are disjoint and there are no hidden atoms. Since
$\sm{\module{P}}=\{\{a\},\{a,b\}\}$ and 
$\sm{\module{Q}}=\{\emptyset,\{a,b\}\}$, we get
$\sm{\module{P}\oplus\module{Q}}=\{\{a,b\}\}
=\sm{\module{P}}\Join\sm{\module{Q}}$. \hfill $\bbox$
\end{example}
Example \ref{ex3} suggests that the denial of positive recursion
between modules can be relaxed in certain cases.   
We define a semantical characterization for module composition that
maintains the compositionality of the stable model semantics.  

\begin{definition}
\label{semantical-join}
The \emph{semantical join} $\module{P}_1\sjoin\module{P}_2$
of two \system{smodels} program modules $\module{P}_1$ and
$\module{P}_2$ is $\module{P}_1\oplus\module{P}_2$, provided
$\module{P}_1\oplus\module{P}_2$ is defined and 
$\sm{\module{P}_1 \oplus
\module{P}_2}=\sm{\module{P}_1}\Join\sm{\module{P}_2}$.
\end{definition}
The module theorem holds by definition for \system{smodels} program
modules composed with $\sjoin$. 
We can now present an alternative formulation for modular equivalence 
taking features from strong equivalence~\cite{LPV01:acmtocl}.

\begin{definition}
\label{new-mod-eq}
\system{smodels} program modules $\module{P}$ and $\module{Q}$ are
\emph{semantically modularly equivalent}, denoted by
$\module{P}\lpeq{sem}\module{Q}$, if 
and only if $\hbi{\module{P}}=\hbi{\module{Q}}$ and
$\module{P}\sjoin\module{R}\lpeq{v}\module{Q}\sjoin\module{R}$ 
for all $\module{R}$ such that $\module{P}\sjoin\module{R}$ and
$\module{Q}\sjoin\module{R}$ are defined.
\end{definition}
It is straightforward to see that $\lpeq{sem}$ is a congruence
for $\sjoin$ and reduces to $\lpeq{v}$ for modules with completely
specified input, that is, modules $\module{P}$ such that
$\hbi{\module{P}}=\emptyset$.  

\begin{theorem}
\label{same-eqs}
$\module{P}\lpeq{m}\module{Q}$ if and only if
$\module{P}\lpeq{sem}\module{Q}$ for any \system{smodels} program
modules $\module{P}$ and $\module{Q}$. 
\end{theorem}

\begin{proof}
Assume  $\module{P}\lpeq{sem}\module{Q}$. Now,
$\module{P}\lpeq{m}\module{Q}$ is implied by Definition
\ref{new-mod-eq} with empty context module
$\module{R}=\tuple{\emptyset,\emptyset,\emptyset,\emptyset}$.   
Assume then $\module{P}\lpeq{m}\module{Q}$, that is, there is a
bijection $f: \sm{\module{P}}\rightarrow\sm{\module{Q}}$ such that for
each $M\in\sm{\module{P}}$,
$M\isect\hbv{\module{P}}=f(M)\isect\hbv{\module{Q}}$.  
Consider arbitrary $\module{R}$ such that
$\module{P}\sjoin\module{R}$ and $\module{Q}\sjoin\module{R}$ are
defined.
Then $\sm{\module{P}\sjoin \module{R}}=
\sm{\module{P}}\Join\sm{\module{R}}$ and
$\sm{\module{Q}\sjoin \module{R}}=
\sm{\module{Q}}\Join\sm{\module{R}}$.

We now define $g: \sm{\module{P}\sjoin \module{R}}\rightarrow
\sm{\module{Q}\sjoin \module{R}}$ 
such that for any
$M\in\sm{\module{P}\sjoin \module{R}}$,
$$g(M)=f(M_P)\union M_R,$$
where $M=M_P\union M_R$ such that $M_P\in\sm{\module{P}}$ and
$M_R\in\sm{\module{R}}$ are compatible.
Now, $g$ is a bijection and 
$M\isect(\hbv{\module{P}}\union\hbv{\module{R}})=
g(M)\isect(\hbv{\module{Q}}\union\hbv{\module{R}})$ for each
$M\in\sm{\module{P}\sjoin\module{R}}$.
Since $\module{R}$ was arbitrary, 
$\module{P}\lpeq{sem}\module{Q}$ follows. 
\end{proof}

Theorem \ref{same-eqs} implies that $\lpeq{m}$ is a congruence for
$\sjoin$, too.
Thus it is possible to replace $\module{P}$ with modularly equivalent
$\module{Q}$ in the contexts allowed by $\sjoin$. 

The syntactical restriction denying positive recursion between modules
is easy to check, since SCCs can be found in a linear time with
respect to the size of the dependency graph~\cite{Tarjan}. 
To the contrary, checking whether $\sm{\module{P}_1\oplus \module{P}_2}=
\sm{\module{P}_1}\Join\sm{\module{P}_2}$ is a computationally harder
problem. 

\begin{theorem}
\label{tradeoff}
Given \system{smodels} program modules $\module{P}_1$ and
$\module{P}_2$ such that $\module{P}_1\oplus\module{P}_2$ is defined,
deciding whether it holds that $\sm{\module{P}_1\oplus \module{P}_2}=
\sm{\module{P}_1}\Join\sm{\module{P}_2}$ is a
$\mathbf{coNP}$-complete decision problem. 
\end{theorem}

\begin{proof}
Let $\module{P}_1$ and $\module{P}_2$ be \system{smodels} program
modules such that $\module{P}_1\oplus\module{P}_2$ is defined.
We can show $\sm{\module{P}_1 \oplus \module{P}_2}\ne
\sm{\module{P}_1}\Join\sm{\module{P}_2}$ by
choosing $M\subseteq\hb{\module{P}_1\oplus\module{P}_2}$ and checking
that  
\begin{itemize}
\item
$M\in\sm{\module{P}_1\oplus\module{P}_2}$ and
$M\isect\hb{\module{P}_1}\not\in\sm{\module{P}_1}$; or
\item
$M\in\sm{\module{P}_1\oplus\module{P}_2}$ and
$M\isect\hb{\module{P}_2}\not\in\sm{\module{P}_2}$; or
\item
$M\not\in\sm{\module{P}_1\oplus\module{P}_2}$,
$M\isect\hb{\module{P}_1}\in\sm{\module{P}_1}$, and
$M\isect\hb{\module{P}_2}\in\sm{\module{P}_2}$.
\end{itemize}
Once we have chosen $M$, these tests can be performed in polynomial
time, which shows that the problem is in $\mathbf{coNP}$. 
To establish  $\mathbf{coNP}$-hardness we present a reduction from
$\overline{\mathbf{3SAT}}$. 
Consider a finite set $S=\eset{C_1}{C_n}$ of three-literal
clauses $C_i$ of the form $l_1\lor l_2\lor l_3$ where each $l_i$ is
either an atom $a$ or its classical negation $\neg a$. 
Each clause $C_i$ is translated into rules $r_{i,j}$ of the form
$c_i\IF f_j$, where 
$1\leq j\leq 3$, and $f_j=a$ if $l_j=a$ and  $f_j=\naf a$ if $l_j=\neg
a$. The intuitive reading of $c_i$ is that clause $C_i$ is satisfied.
We define modules
$\module{P}_1=\tuple{\{e\IF d.\}, \{d\}, \{e\},\emptyset}$ and
\begin{multline*}
\module{P}_2=\langle 
\{r_{i,j}\mid 1\leq i\leq n,1\leq j\leq 3\}\union \{d\IF
e,c_1,\ldots, c_n.\}, \\ \hb{S}\union\{e\}, \{d\},\{c_1,\ldots,c_n\}
\rangle.  
\end{multline*}
Now $\module{P}_1\oplus\module{P}_2$ is defined, and
$\sm{\module{P}_1}=\{\emptyset,\{d,e\}\}$.
There is $M\in\sm{\module{P}_2}$ that is compatible with $\{d,e\}$
if and only if $S\in\mathbf{3SAT}$.
Since $d\not\in N$ and $e\not\in N$ for all
$N\in\sm{\module{P}_1\oplus\module{P}_2}$, it follows that 
$S\in\overline{\mathbf{3SAT}}$ if and only if
$\sm{\module{P}_1\oplus\module{P}_2}=
\sm{\module{P}_1}\Join\sm{\module{P}_2}$.  
\end{proof}
Theorem \ref{tradeoff} shows that there is a tradeoff for allowing
positive recursion between modules, as more effort is needed to check
that composition of such modules does not compromise the
compositionality of the stable model semantics.

%------------------------------------------------------------------------------

\section{Tools and Practical Demonstration}
\label{section:experiments}

The goal of this section is to demonstrate how the module system
introduced in Section \ref{section:modules} can be exploited in
practise in the context of the \system{smodels} system and other
compatible systems. In this respect, we present tools that have been
developed for the (de)composition of logic programs that are
represented in the
\emph{internal file format}%
\footnote{The reader is referred to \cite{Janhunen07:sea}
for a detailed description and analysis of the format.}
of the \system{smodels} engine.  The binaries for both tools 
are available under the \system{asptools} collection% 
\footnote{\url{http://www.tcs.hut.fi/Software/asptools/}}.
Moreover, we conduct and report a practical experiment which
illustrates the performance of the tools when processing substantially 
large benchmark instances, that is, \system{smodels} programs having
up to millions of rules (see the \system{asptools} web page for
examples). 

The first tool, namely \system{modlist}, is targeted at program
decomposition based on the strongly connected components of an
\system{smodels} program given as input. In view of the objectives
of Section \ref{section:decomposition}, there are three optional
outcomes of the decomposition, that is, strongly connected components
that take into account
\begin{enumerate}
\item positive dependencies only,
\item positive dependencies and hidden atoms, and
\item both positive and negative dependencies as well as hidden atoms.
\end{enumerate}
The number of modules created by \system{modlist} decreases in this
order.  However, our benchmarks cover program instances that get split
in tens of thousands of modules. To tackle the problem of storing such
numbers of modules in separate files we decided to use file compression
and packaging tools and, in particular, the \system{zip} utility
available in standard Linux installations. We found \system{zip}
superior to \system{tar} as it allows random access to files in
an archive, or a \emph{zipfile}. This feature becomes valuable
when the modules are accessed from the archive for further
processing.

The tool for program composition has been named as \system{lpcat}
which refers to the concatenation of files containing logic programs.
A new version of the tool was implemented for experiments reported
below for better performance as well as usability. The old version
(version 1.8) is only able to combine two modules at a time which
gives a quadratic nature for a process of combining $n$ modules
together: modules are added one-by-one to the composition. The new
version, however, is able to read in modules from several files and,
even more conveniently, a \emph{stream of modules} from an individual
file. The \system{zip} facility provides an option for creating such a
stream that can then be forwarded for \system{lpcat} for composition.
This is the strategy for composing programs in experiments that
are described next.

\begin{table}
\begin{center}
\begin{tabular}{@{\hspace{-3pt}}c@{\hspace{-3pt}}}
\begin{tabular}{lrrlrrr}
\textbf{Benchmark} &
\textbf{na} & \textbf{nr} & \textbf{bt} &
\textbf{nm} & \textbf{dt} (s) & \textbf{ct} (s) \\
\hline \hline
ephp-13          & 35 518 &     90 784 & $+$  &  35 518 &  2 110 &    362 \\
                 &        &            & $+$h &  35 518 &  2 110 &    362 \\
                 &        &            &$\pm$h&  35 362 &  2 090 &    361 \\
\hline
mutex3           & 276 086 & 2 406 357 & $+$  & 101 819 & 22 900 &  9 570 \\
                 &         &           & $+$h & 101 819 & 23 300 &  9 640 \\
                 &         &           &$\pm$h& 101 609 & 24 000 &  9 580 \\
\hline
phi3             &   7 379 &    14 274 & $+$  &   6 217 &   74,3 &   3,32 \\
                 &         &           & $+$h &   6 217 &   74,3 &   3,35 \\
                 &         &           &$\pm$h&   5 686 &   63,2 &   2,92 \\
\hline
seq4-ss4    &   6 873 & 1 197 182      & $+$  &   3 425 &   121 & 60,0  \\
                 &         &           & $+$h &   1 403 &   89,4 & 31,9 \\
                 &         &           &$\pm$h &    107 &   20,2 &  7,58 \\
\hline
\end{tabular} \\
\ \\
\begin{tabular}{@{}lll@{}}
Legends for abbreviations:
& \textbf{na}: & Number of atoms \\
& \textbf{nr}: & Number of rules \\
& \textbf{nm}: & Number of modules \\
& \textbf{bt}: & Benchmark type \\
& \textbf{dt}: & Decomposition time \\
& \textbf{ct}: & Composition time \\
\end{tabular}
\end{tabular}
\end{center}
\caption{Summary of benchmark results for module (de)composition
         \label{table:results}}
\end{table}

To test the performance of our tools, we picked a set of benchmark
instances having from tens of thousands up to millions of
rules---expressed in the \system{smodels} format.
For each instance, the first task is to decompose the instance into
modules using \system{modlist} and to create a zipfile containing the
modules. The type of modules to be created is varied according the
three schemes summarized above.
The second task is to recreate the benchmark instance from a stream of
modules extracted from the respective zipfile. As suggested above, the
actual composition is carried out using \system{lpcat} and we also
check that the number of rules matches with the original instance.
Due to high number of rules, checking the equivalence of the original
and composed programs \cite{JO07:tplp} is unfeasible in many cases. If
all atoms are visible, this can be accomplished syntactically on the
basis of sorted textual representations of the programs involved.  To
ensure that \system{modlist} and \system{lpcat} produce correct
(de)compositions of programs, such a check was performed for all
compositions created for the first three benchmarks which involve no
hidden atoms.
As regards computer hardware, we run \system{modlist} and
\system{lpcat} on a PC with a 1.8GHz Intel Core 2 Duo CPU and 2GBs of
main memory---operating under the Linux 2.6.18 system. In experimental
results collected in Table \ref{table:results}, we report the sum of
user and system times that are measured with the {\tt /usr/bin/time}
command. There are three benchmark types (\textbf{bt} for short)
as enumerated in the beginning of this section. We refer to them
using the respective abbreviations $+$, $+$h, and $\pm$h.

The first benchmark instance in Table \ref{table:results},
viz.~\emph{ephp-13}, is a formalization \cite{JO07:iclp} of the
classical \emph{pigeon hole principle} for 13 pigeons---extended by
redundant rules in analogy to Tseitin's \emph{extended resolution}
proof system. This program can be deemed medium-sized within our
benchmarks. There are no hidden atoms, no positive recursion and
little negative recursion in this program instance as indicated by the
number of atoms (35 518) and the respective numbers of modules (see
column \textbf{nm}). Thus we have an example of a very fine-grained
decomposition where the \emph{definition}%
\footnote{The set of rules that mention the atom in question in their head.}
of each atom ends up as its own module in the outcome. The given
timings indicate that \system{modlist} and \system{lpcat} are able to
handle $15$ and $100$ modules per second, respectively. The share
of file I/O and (de)compression is substantial in program decomposition.
For instance, the actual splitting of the \emph{ephp-13} benchmark
($+$h) using \system{modlist} takes only $0,59$ seconds---the
rest of approximately~$2 110$ seconds is spent to create the zipfile.
To the contrary, inflating the stream of modules from the zipfile is
very efficient as it takes only $0,45$ seconds in case of
\emph{ephp-13}.  After that the restoration of the original program
instance takes roughly $361$ seconds. The creation and compression of
a joint symbol table for the modules accounts for the most of the time
spent on this operation.
It should also be stressed that it is impractical to store modules in
separate files for this program. For instance, a shell command that
refers to all modules fails due to excessive number of arguments
at the respective command line.

The next two programs in Table \ref{table:results}, \emph{mutex3} and
\emph{phi3}, are related to the \emph{distributed implementability
  problem} of asynchronous automata, and particular formalizations of
classical \emph{mutual exclusion} and \emph{dining philosophers}
problems \cite{HS04:report,HS05:acsd}. These programs involve no
hidden atoms and both positive and negative interdependencies of atoms
occur. The extremely high numbers of rules ($2 406 357$) and modules
($101 819$) are clearly reflected in running times perceived for
\emph{mutex3}. However, the respective rates of $4$ and $10$ modules
per second do not differ too much from those obtained for
\emph{ephp-13} given the fact that the number of rules is about 25
times higher.
The data observed for \emph{phi3} is analogous to those obtained for
\emph{ephp-13} and \emph{mutex3} but the respective modules-per-second
rates are much higher: approximately~$90$ and $2000$. This may partly boil
down to the fact \emph{phi3} is the smallest program under
consideration and it has also the smallest number of rules
per module ratio.

Our last benchmark program, \emph{seq4-ss4}, is taken from
benchmark sets of \cite{BCVF06:iclp} where the optimization of machine
code using ASP techniques is of interest. The program in question
formalizes the optimization of a particular sequence of four SPARC-v7
instructions. This program instance has the greatest modules as
regards the number of rules---the average number of rules per module
varies from about $350$ to $11 200$ depending on the module type. It
has also hidden atoms which makes a difference between modules based
on plain SCCs and their combinations induced by the dependencies
caused by the use of hidden atoms. The respective modules-per-second
rates $28$, $19$, and $5$ are all better than $4$ obtained for
\emph{mutex3}.

To provide the reader with a better idea of sizes of individual modules,
we have collected some numbers about their distribution in Table
\ref{table:distribution}. Each program involves a substantial
number of modules with just one rule each of which defines
a single atom of interest.
On the other hand, the largest SCCs for \emph{ephp-13}, \emph{mutex3},
\emph{phi3}, and \emph{seq4-ss4} involve $949$, $2 091 912$, $2 579$,
and $1071689$ rules, respectively. For \emph{mutex3}, the biggest
module consists of a definition of an equivalence relation over states
in the verification domain---creating a huge set of positively
interdependent atoms. For \emph{ephp-13}, the greatest module is a
collection of nogoods which can be shown to have no stable models in
roughly $99500$ seconds using \system{smodels} (version 2.32).
However, the remaining rules of \emph{ephp-13} make this fact much
faster to prove: only 61 seconds elapse.

\begin{table}
\begin{center}
\begin{tabular}{lrrrr}
\hline
Benchmark    & ephp-13 & mutex3 & phi3 & seq-ss4 \\
\hline \hline
\textbf{nr}  & \multicolumn{4}{c}{\textbf{nm}} \\
\hline 
1            & 14 474  & 67 749 & 2 811 & 2 969 \\
2            &  7 014  &  2 757 & 1 434 &       \\
3--4         & 12 680  &     41 & 1 962 &       \\
5--8         &    149  & 30 798 &     2 &       \\
9--16        &    618  &    255 &     6 &       \\
17--32       &    582  &        &       &    11 \\
33--64       &         &        &     1 &       \\
65--128      &         &        &       &   134 \\
129--512     &         &        &       &   296 \\
513--1 024   &      1  &        &       &     9 \\
over 1 024   &         &      2 &     1 &     2 \\
\hline
\end{tabular}
\end{center}
\caption{Distribution of the sizes of modules \label{table:distribution}
         (see Table \ref{table:results} for legends)}
\end{table}

A few concluding remarks follow.
Increasing the number of modules in a program tends to decrease the
number of modules that can be decomposed per time unit. This
observation suggests that the creation of the zipfile has a quadratic
flavor although modules themselves can be figured out in linear time
(using a variant of Tarjan's algorithm). Perhaps this can be improved
in the future by better integrating the creation of the zipfile into
\system{modlist}. For now, it creates a shell script for this purpose.
Handling the biggest program instances is also subject to the effects
of memory allocation which may further slow down computations.
On the other hand, the cost of increasing the number of rules in modules
seems to be relatively small.
Moreover, it is clear on the basis of data given in Table
\ref{table:results} that the composition of programs is faster than
decomposition. This would not be the case if the old version
of \system{lpcat} were used for composition.
Last, we want to emphasize that \system{modlist} and \system{lpcat}
have been implemented as supplementary tools that are not directly
related to the computation of stable models. Nevertheless, we intend
to exploit these tools in order to modularize different tasks in ASP
such as verifying ordinary/modular equivalence and program
optimization. The existence of such tools enables modular program
development and the creation of \emph{module libraries} for
\system{smodels} programs, and thus puts forward the use of module
architectures in the realm of ASP.

%------------------------------------------------------------------------------

\section{Conclusions}
\label{section:conclusions}

In this paper, we introduce a simple and intuitive notion of a logic
program module that interacts with other modules through a
well-defined input/output interface. The design has its roots in a
module architecture proposed for conventional logic programs
\cite{GS89}, but as regards our contribution, we tailor the
architecture in order to better meet the criteria of ASP. Perhaps the
most important objective in this respect is to achieve the
compositionality of stable model semantics, that is, the semantics of
an entire program depends directly on the semantics assigned to its
modules.
To this end, the main result of this paper is formalized as the
\emph{module theorem} (Theorem \ref{moduletheorem}) which links
program-level stability with module-level stability. The theorem holds
under the assumption that positively interdependent atoms are always
placed in the same module. The \emph{join} operation $\join$ defined
for program modules effectively formalizes this constraint---which we
find acceptable when it comes to good programming style in ASP.

The module theorem is also a proper generalization of the
splitting-set theorem~\cite{LT94:iclp} recast for \system{smodels}
programs. The main difference is that splitting-sets do not enable any
kind of recursion between modules. Even though the module theorem is
proved to demonstrate the feasibility of the respective module
architecture, it is also applied as a tool to simplify mathematical
proofs in this paper and recently also in~\cite{OJ08:aimsa,OJ08:jlc}.   
It also lends itself to extensions for further classes of logic
programs which can be brought into effect in terms of \emph{strongly
  faithful}, $\join$-\emph{preserving}, and \emph{modular}
translations for the removal of new syntax (Theorem
\ref{theorem:modulethr-translation}). Moreover, the module theorem
paves the way for the modularization of various reasoning tasks, such 
as search for answer sets, query evaluation, and verification, in ASP.

The second main theme of the paper is the notion of modular
equivalence which is proved to be a proper congruence relation for
program composition using $\join$ (Theorem \ref{smodels-congruence}).
Thus modular equivalence is preserved under substitutions of modularly
equivalent program modules. Since uniform equivalence is not a
congruence for ordinary $\union$ but strong equivalence is by
definition, modular equivalence can be viewed as a reasonable
compromise between these two extremes. In addition to the congruence
property, we present a number of results about modular equivalence.
\begin{enumerate}
\item
We show that deciding modular equivalence forms a
$\mathbf{coNP}$-complete decision problem for \system{smodels}
program modules with the EVA property, that is, those having enough
visible atoms so that their stable models can be distinguished from
each other on the basis of visible atoms only. In this way, it is
possible to use the \system{smodels} solver for the actual
verification task.

\item
We consider the possibility of redefining the join operation $\join$
using a semantical condition that corresponds to the content of the
module theorem. The notion of modular equivalence is not affected,
but the cost of verifying whether a particular join of modules is
defined becomes a $\mathbf{coNP}$-complete decision problem.
This is in contrast with the linear time check for positive recursion
(Tarjan's algorithm for strongly connected components) but it may
favorably extend the coverage of modular equivalence in certain
applications.

\item
Finally, we also analyze the problem of decomposing an
\system{smodels} program into modules when there is no a priori
knowledge about the structure of the program. The strongly connected
components of the program provide the starting point in this
respect, but the usage of hidden atoms may enforce a higher degree
of amalgamation when the modules of a program are extracted.
\end{enumerate}
The theoretical results presented in the paper have emerged in close
connection with the development of tools for ASP. The practical
demonstration in Section \ref{section:experiments} illustrates the
basic facilities that are required to deal with \emph{object level}
modules within the \system{smodels} system.%
\footnote{Likewise, source level modules could be incorporated
to the front-end of the system (\system{lparse}).}
The linker, namely \system{lpcat}, enables the composition of ground
programs in the \system{smodels} format. Using this tool, for instance,
it is possible to add a query to a program afterwards without grounding
the program again.
On the other hand, individual modules of a program can be accessed
from the zipfile created by the module extractor \system{modlist}.
This is highly practical since we intend to pursue techniques for
module-level optimization in the future.

%------------------------------------------------------------------------------

\section*{Acknowledgements}

This work has been partially supported by the Academy of Finland
through Projects \#211025 and \#122399. 
The first author gratefully acknowledges the financial support from
Helsinki Graduate School in Computer Science and Engineering, Emil
Aaltonen Foundation, the Finnish Foundation for Technology
Promotion TES, the Nokia Foundation, and the Finnish Cultural
Foundation.  

\appendix

%------------------------------------------------------------------------------

\section{Proofs}
\label{proofs}

\begin{proof}[Proof of Theorem \ref{smodels-congruence}]
Let $\module{P}$ and $\module{Q}$ be modules such that
$\module{P}\lpeq{m}\module{Q}$. 
Let $\module{R}$ be an arbitrary
module such that $\module{P}\join \module{R}$ and $\module{Q}\join
\module{R}$ are defined.   
From $\hbv{\module{P}}=\hbv{\module{Q}}$ and
$\hbi{\module{P}}=\hbi{\module{Q}}$ it follows 
that
$\hbv{\module{P}\join\module{R}}=\hbv{\module{Q}\join\module{R}}$ and
$\hbi{\module{P}\join\module{R}}=\hbi{\module{Q}\join\module{R}}$.

Consider any model $M\in\sm{\module{P}\join\module{R}}$.
By Theorem \ref{moduletheorem},
$\sm{\module{P}\join\module{R}}=\sm{\module{P}}\Join\sm{\module{R}}$,
that is, 
$M_P=M\isect\hb{\module{P}}\in\sm{\module{P}}$ 
and
$M_R=M\isect\hb{\module{R}}\in\sm{\module{R}}$.
Since $\module{P}\lpeq{m}\module{Q}$, there is a bijection $f:
\sm{\module{P}}\rightarrow\sm{\module{Q}}$ such that
$M_P\in\sm{\module{P}} 
\Longleftrightarrow f(M_P)\in\sm{\module{Q}}$, and
\begin{equation}
M_P\isect \hbv{\module{P}}=f(M_P)\isect \hbv{\module{Q}}.
\label{eq4}
\end{equation} 
Denote $M_Q=f(M_P)$. Clearly, $M_P$ and $M_R$ are compatible. Since
(\ref{eq4}) holds, also $M_Q$ and $M_R$ are compatible. Applying
Theorem \ref{moduletheorem} we get $M_Q\union
M_R\in\sm{\module{Q}\join\module{R}}=\sm{\module{Q}}\Join\sm{\module{R}}$.  

Now, define a function $g:\sm{\module{P}\join\module{R}}
\rightarrow \sm{\module{Q}\join\module{R}}$ as
$$g(M)=f(M\isect\hb{\module{P}})\union (M\isect\hb{\module{R}}).$$ 
Clearly, $g$ maps the set of visible atoms in $M$ to itself, that is,
$$M\isect (\hbv{\module{P}\join\module{R}})  
=g(M)\isect (\hbv{\module{Q}\join\module{R}}).$$
Function $g$ is a bijection, since 
\begin{itemize}
\item $g$ is an injection: $M\ne N$ implies
$g(M)\ne g(N)$ for all $M,N\in\sm{\module{P}\join\module{R}}$, since 
$f(M\isect\hb{\module{P}})\ne f(N\isect\hb{\module{P}})$ or
$M\isect\hb{\module{R}}\ne N\isect\hb{\module{R}}$.
\item $g$ is a surjection: for any
$M\in\sm{\module{Q}\join\module{R}}$,
$N=f^{-1}(M\isect\hb{\module{Q}})\union(M\isect\hb{\module{R}})\in
\sm{\module{P}\join\module{R}}$ and $g(N)=M$, since $f$ is a
surjection. 
\end{itemize}
The inverse function $g^{-1}:\sm{\module{Q}\join\module{R}}\rightarrow 
\sm{\module{P}\join\module{R}}$ can be defined as 
$g^{-1}(N)=f^{-1}(N\isect\hb{\module{Q}})\union
(N\isect\hb{\module{R}})$. 
Thus $\module{P}\join \module{R} \lpeq{m} \module{Q}\join \module{R}$. 
\end{proof}

\begin{proof}[Proof of Theorem \ref{theorem:nlp-moduletheorem}]
We present an alternative proof to the one given in \cite{OJ06:ecai}. 
We use the characterization of stable models based on the programs
completion and loop formulas presented in
Theorem~\ref{stable-models-using-loop-formulas}.

First, we need to generalize the concepts of completion and
loop formulas for NLP modules. 
Given a normal logic program module $\module{P}=\tuple{R,I,O,H}$,
we define 
\begin{eqnarray}
\clark{\module{P}}=
\bigwedge_{a \in O\union H}
\Bigg (a \leftrightarrow
\bigvee_{\head{r}=a} 
\Bigg ( \bigwedge_{b \in \pbody{r}} b  \wedge
\bigwedge_{c \in \nbody{r}} \neg c \Bigg ) \Bigg), 
\end{eqnarray}
that is, we take the completion in the normal way for the set of rules
$R$ with the exception that we take into account that input
atoms do not have any defining rules. 
As regards loop formulas, we define $\lfs{\module{P}}=\lfs{R}$,
since no atom in the input signature can appear in any of the loops. 

Consider an arbitrary NLP module $\module{P}=\tuple{R,I,O,H}$. Define
a set of rules, that is, a conventional normal logic program, $R'=R\union
G_I$, where $G_I=\set{a\IF \naf \bar a\rsep \bar a\IF\naf 
a\mid a\in I}$ and all atoms $\bar a$ are new atoms not appearing in
$\hb{\module{P}}$.  
Now, 
$M\in\sm{\module{P}}$ if and only if $N=M\union\set{\bar a\mid a\in
I\setminus M}\in\sm{R'}$.
On the other hand, by Theorem \ref{stable-models-using-loop-formulas},
$N\in\sm{R'}$ if and only if $N\models\clark{R'}\union\lfs{R'}$. 
Considering the completion, since $\head{R}\isect I=\emptyset$, and 
atoms $\bar a$ are new, it holds that
$\clark{R'}=\clark{\module{P}}\union\clark{G_I}$. 
As regards loop formulas, we notice that $\dep{R'}$ is $\dep{R}$
together with vertices for atoms $\set{\bar a\mid a\in I}$, which have
no edges in $\dep{R'}$.
Therefore, 
$\lfs{R'}=\lfs{R}=\lfs{\module{P}}$.

Thus $N\models\clark{R'}\union\lfs{R'}$ if and only if
$N\models\clark{\module{P}}\union\clark{G_I}\union\lfs{\module{P}}$.
Furthermore, based on the relationship between $M$ and $N$, it holds
that
$N\models\clark{\module{P}}\union\clark{G_I}\union\lfs{\module{P}}$ if 
and only if $M\models\clark{\module{P}}\union\lfs{\module{P}}$.
Thus Theorem \ref{stable-models-using-loop-formulas} generalizes
directly for NLP modules:
given a normal logic program module $\module{P}$ and an interpretation 
$M\subseteq\hb{\module{P}}$, it holds that $M\in\sm{\module{P}}$ if
and only if $M\models\clark{\module{P}}\union\lfs{\module{P}}$.

Now, since the join operation does not allow positive recursion
between two modules, and
$(\hbo{\module{P}_1}\union\hbh{\module{P}_1})\isect(\hbo{\module{P}_2} 
\union\hbh{\module{P}_2})=\emptyset$, it holds that
\begin{eqnarray}
\clark{\module{P}_1\join\module{P}_2} &=&\clark{\module{P}_1}\union
\clark{\module{P}_2},\mbox{ and }\\
\lfs{\module{P}_1\join\module{P}_2} &=&\lfs{\module{P}_1}\union
\lfs{\module{P}_2}.
\end{eqnarray}

Furthermore, the satisfaction relation is compositional for $\union$,
that is, $M\models P\union Q$ if and only if
$M\isect\hb{P}\models P$ and $M\isect\hb{Q}\models Q$ for any
propositional theories $P$ and $Q$. Thus
\begin{eqnarray*}
M\in\sm{\module{P}_1\join\module{P}_2} 
& \Longleftrightarrow &
M\models\clark{\module{P}_1\join\module{P}_2}\union
\lfs{\module{P}_1\join\module{P}_2}\\
& \Longleftrightarrow &
M\models\clark{\module{P}_1}\union\clark{\module{P}_2}\union
\lfs{\module{P}_1}\union\lfs{\module{P}_2}\\
& \Longleftrightarrow &
M\isect\hb{\module{P}_1}\models\clark{\module{P}_1}\union
\lfs{\module{P}_1}  \mbox{ and} \\ &&
M\isect\hb{\module{P}_2}\models\clark{\module{P}_2}\union
\lfs{\module{P}_2}\\
& \Longleftrightarrow &
M\isect\hb{\module{P}_1}\in\sm{\module{P}_1}\mbox{ and }
M\isect\hb{\module{P}_2}\in\sm{\module{P}_2}.
\end{eqnarray*}
It follows that $\sm{\module{P}_1\join\module{P}_2}=
\sm{\module{P}_1}\Join\sm{\module{P}_2}$.
\end{proof}

\begin{proof}[Proof of Theorem \ref{theorem:modulethr-translation}]
Let $\mathcal{C}_1$ and $\mathcal{C}_2$ be two classes of logic
program modules such that $\mathcal{C}_2\subseteq\mathcal{C}_1$, 
and the module theorem holds for modules in $\mathcal{C}_2$.
Consider a translation function $\trop{}\!: \mathcal{C}_1\rightarrow
\mathcal{C}_2$ such that Conditions 1--3 from Definition
\ref{def:conditions-for-translation} are satisfied.
Let $\module{P}_1,\module{P}_2\in \mathcal{C}_1$ be  
modules such that $\module{P}_1\join\module{P}_2$ is defined. 
Then Condition 2 implies that
$\tr{}{\module{P}_1},\tr{}{\module{P}_2}\in\mathcal{C}_2$ are 
modules such that $\tr{}{\module{P}_1}\join\tr{}{\module{P}_2}$ is
defined. 
Since the module theorem holds for modules in $\mathcal{C}_2$,
$\sm{\tr{}{\module{P}_1}\join\tr{}{\module{P}_2}}=
\sm{\tr{}{\module{P}_1}}\Join\sm{\tr{}{\module{P}_2}}$.
Moreover, Condition 3 implies that 
$\sm{\tr{}{\module{P}_1\join\module{P}_2}}
=\sm{\tr{}{\module{P}_1}\join\tr{}{\module{P}_2}}$.

Condition 1 implies there is a bijection
\begin{eqnarray*}
g:&& \sm{\module{P}_1\join\module{P}_2}\rightarrow
\sm{\tr{}{\module{P}_1\join\module{P}_2}}
\end{eqnarray*}
such that for any $M\in\sm{\module{P}_1\join\module{P}_2}$ we have
$M=g(M)\isect\hb{\module{P}_1\join\module{P}_2}$. 
Notice that strong faithfulness requires that the projections of $M$
and $g(M)$ have to be identical over whole
$\hb{\module{P}_1\join\module{P}_2}$ not just over
$\hbv{\module{P}_1\join\module{P}_2}$. 
Similarly there are bijections
\begin{eqnarray*}
g_1:&& \sm{\module{P}_1}\rightarrow \sm{\tr{}{\module{P}_1}}
\mbox{ and }\\
g_2: &&\sm{\module{P}_2}\rightarrow \sm{\tr{}{\module{P}_2}}
\end{eqnarray*}
such that for any $M_i\in\sm{\module{P}_i}$ ($i=1,2$) it holds
that $M_i=g_i(M_i)\isect\hb{\module{P}_i}$.
Consider arbitrary $M\subseteq \hb{\module{P}_1\join\module{P}_2}$,
and its projections $M_1=M\isect\hb{\module{P}_1}$ and
$M_2=M\isect\hb{\module{P}_2}$. Now,
$M_1$ and $M_2$ are compatible, and $M=M_1\union M_2$.

Assume that $M\in\sm{\module{P}_1\join\module{P}_2}$. 
Since the module theorem holds for $\mathcal{C}_2$, we have
$$g(M)\in \sm{\tr{}{\module{P}_1\join\module{P}_2}}=
\sm{\tr{}{\module{P}_1}\join\tr{}{\module{P}_2}}=
\sm{\tr{}{\module{P}_1}}\Join\sm{\tr{}{\module{P}_2}},$$ 
that is,\vspace{-1ex}
\begin{eqnarray*}
N_1&=&g(M)\isect\hb{\tr{}{\module{P}_1}}\in\sm{\tr{}{\module{P}_1}},
\\
N_2&=&g(M)\isect\hb{\tr{}{\module{P}_2}}\in\sm{\tr{}{\module{P}_2}},
\mbox{ and}  
\end{eqnarray*}
$N_1$ and $N_2$ are compatible projections of $g(M)$.
Moreover, 
$M=g(M)\isect\hb{\module{P}_1\join\module{P}_2}$ and
$M_i=N_i\isect\hb{\module{P}_i}$ for $i=1,2$.
Using the inverse functions of $g_1$ and $g_2$ we get
$M_1= g_1^{-1}(N_1)\in\sm{\module{P}_1}$ and 
$M_2=g_2^{-1}(N_2)\in\sm{\module{P}_2}$.

For the other direction, assume that
$M_1\in\sm{\module{P}_1}$ and  
$M_2\in\sm{\module{P}_2}$. Then $N_1=g_1(M_1)\in\sm{\tr{}{\module{P}_1}}$
and $N_2=g_2(M_2)\in\sm{\tr{}{\module{P}_2}}$. 
Since 
\begin{eqnarray*}
N_1\isect\hbv{\tr{}{\module{P}_2}}&=&M_1\isect\hbv{\module{P}_2},\\
N_2\isect\hbv{\tr{}{\module{P}_1}}&=&M_2\isect\hbv{\module{P}_1},\mbox{ and}
\end{eqnarray*}
$M_1$ and $M_2$ are compatible, also $N_1$ and
$N_2$ are compatible.
By applying the module theorem for
$\mathcal{C}_2$, we get
$N=N_1\union N_2\in\sm{\tr{}{\module{P}_1}\join\tr{}{\module{P}_2}}=
\sm{\tr{}{\module{P}_1\join\module{P}_2}}$.
Furthermore, $N\isect\hb{\module{P}_1\join\module{P}_2}=M_1\union
M_2=M$, and using the inverse of $g$ we get
$M=g^{-1}(N)\in\sm{\module{P}_1\join\module{P}_2}$.

Thus  we have shown $\sm{\module{P}_1\join\module{P}_2}=
\sm{\module{P}_1}\Join\sm{\module{P}_2}$. %\hfill $\Box$ 
\end{proof}

\begin{proof}[Proof of Theorem \ref{theorem:translation-smodels2nlp}]
Consider \system{smodels} program modules $\module{P}_1$ and
$\module{P}_2$. 
It is straightforward to see that $\trop{NLP}$ is $\join$-preserving,
that is, if $\module{P}_1\join\module{P}_2$ is defined, then also
$\tr{NLP}{\module{P}_1}\join\tr{NLP}{\module{P}_2}$ is defined. 
The key observation is that for every edge in the dependency graph
$\dep{\tr{NLP}{\module{P}_1}\join\tr{NLP}{\module{P}_2}}$
there is also an edge in
$\dep{\module{P}_1\join\module{P}_2}$.\footnote{It might be the case
  that $\dep{\module{P}_1\join\module{P}_2}$ contains some edges that
  are not in $\dep{\tr{NLP}{\module{P}_1}\join\tr{NLP}{\module{P}_2}}$.
  This happens when there is a weight rule the body of which can never
  be satisfied.} 
Since $\module{P}_1$ and $\module{P}_2$ are mutually independent,
also $\tr{NLP}{\module{P}_1}$ and $\tr{NLP}{\module{P}_2}$ are mutually
independent. 
Furthermore, if $\module{P}_1\join\module{P}_2$ is defined, then
$\tr{NLP}{\module{P}_1}$ and $\tr{NLP}{\module{P}_2}$ respect
each other's hidden atoms. This is because new atoms are introduced 
only for $\choiceheads{\module{P}_1}\subseteq\hbo{\module{P}_1}\union
\hbh{\module{P}_1}$ and
$\choiceheads{\module{P}_2}\subseteq\hbo{\module{P}_2}\union 
\hbh{\module{P}_2}$, and 
$(\hbo{\module{P}_1}\union
\hbh{\module{P}_1})\isect(\hbo{\module{P}_2}\union 
\hbh{\module{P}_2})=\emptyset$.
Since 
$$\tr{NLP}{\module{P}_1}\join\tr{NLP}{\module{P}_2}=
\tr{NLP}{\module{P}_1\join\module{P}_2}$$
holds, $\trop{NLP}$ is also modular.
Thus $\trop{NLP}$ satisfies conditions 2 and 3 in
Definition~\ref{def:conditions-for-translation}. 

We are left to show that $$\reveal{\module{P},H}
\lpeq{m} \reveal{\tr{NLP}{\module{P}}, H}$$ for any 
\system{smodels} program module $\module{P}=\tuple{P,I,O,H}$ and
its translation $\tr{NLP}{\module{P}}=\tuple{R,I,O, H\union H'}.$ 
Note that $\sm{\module{P}}=\sm{\reveal{\module{P},H}}$
and  $\sm{\tr{NLP}{\module{P}}}=\sm{\reveal{\tr{NLP}{\module{P}},H}}$,
and the additional restriction imposed by revealing is that the
bijection between these sets of stable models needs to be such
that their projections over $\hb{\module{P}}$, not just
over $\hbv{\module{P}}$, coincide.

We define a function
$f:\sm{\module{P}}\rightarrow 2^{\hb{\tr{NLP}{\module{P}}}}$ such that
$$f(M)=M\union\{\overline{a} \mid a\in\choiceheads{P}\setminus M\}.$$ 
Clearly $M=f(M)\isect\hb{\module{P}}$. We need to show that 
\begin{itemize}
\item[(i)] given any $M\in\sm{\module{P}}$,
$f(M)\in\sm{\tr{NLP}{\module{P}}}$; and
\item[(ii)] $f:\;\sm{\module{P}}\rightarrow \sm{\tr{NLP}{\module{P}}}$
  is a bijection.   
\end{itemize}
Note that for any atom~$a\in\choiceheads{P}$, it holds that
$\overline{a}\in f(M)$ if and only if $a\not\in M$ if and only if
$a\not\in f(M)$. 

\begin{itemize}
\item[(i)] \emph{We show that $N=f(M)\in\sm{\tr{NLP}{\module{P}}}$ for
    any $M\in \sm{\module{P}}$:} 

Assume first $N\not\models\GLred{R}{N}{I}$, that is, there is a rule
$r$ in $\GLred{R}{N}{I}$ that is not satisfied by $N$.
If $\head{r}\in\hb{\module{P}}$, then $r$ is of the form
$a\IF B'\setminus I$ and there are two possibilities:
\begin{enumerate}
\item
There is a rule $a\IF B',\naf
C,\naf\overline{a}\in R$ corresponding to a choice rule 
$\{A\}\IF B', \naf C\in P$ such that $a\in A$, $\overline{a}\not\in
N$, $B'\isect I\subseteq N$, and $C\isect N=\emptyset$.  
Since $\overline{a}\not\in N$ implies $a\in N$, $r$ is satisfied in
$N$, a contradiction. 
\item
There is a rule $a\IF B',\naf C'\in R$
corresponding to a weight rule 
$r'=a\IF\limit{w}{B=W_{B},\naf C=W_{C}}\in P$ such that 
$B'\subseteq B$, $C'\subseteq C$, 
$w\leq \sum_{b\in B'}w_b +\sum_{c\in C'}w_c$,  
$B'\isect I\subseteq N$, and $C'\isect N=\emptyset$. 
Since $N\not\models r$ we must have $a\not\in N$ (which implies
$a\not\in M$) and $B'\setminus I\subseteq N$. Thus $B'\subseteq N$
which implies $B'\subseteq M$. Moreover, $C'\isect N=\emptyset$
implies $C'\isect M=\emptyset$. 
But then $w\leq\sum_{b\in B'\isect M}w_b +\sum_{c\in C'\setminus
  M}w_c$. Since $B'\subseteq B$ and  $C'\subseteq C$, we have
$w\leq\sum_{b\in B\isect M}w_b +\sum_{c\in C\setminus M}w_c$ and
$M\not\models r'$, a contradiction. 
\end{enumerate}
Otherwise each $r\in\GLred{R}{N}{I}$ is of the form $\overline{a}$, in
which case there is a rule 
$\overline{a}\IF \naf a\in R$ and $a\not\in N$. Since 
$a\not\in N$ implies $\overline{a}\in N$, then $N\models r$,
a contradiction. 
Thus $N\models\GLred{R}{N}{I}$, and furthermore $N\setminus
I\models\GLred{R}{N}{I}$.  

\hspace{2ex} Assume now $N\setminus I\ne\lm{\GLred{R}{N}{I}}$, that
is, there is 
$N'\subset N\setminus I$ such that $N'\models\GLred{R}{N}{I}$. 
We define $M'=N'\isect\hb{\module{P}}$ and show $M'\models
\GLred{P}{M}{I}$, which contradicts the assumption $M\setminus
I=\lm{\GLred{P}{M}{I}}$, since $M'\subset M\setminus I$.
Assume that there is a rule $r\in\GLred{P}{M}{I}$ such that
$M'\not\models r$. There are two possibilities: 
\begin{enumerate}
\item
$r$ is of the form $a\IF B\setminus I$. 
Then there is a choice rule $\{A\}\IF B, \naf C\in P$, such that
$B\isect I\subseteq M$, $C\isect M=\emptyset$, and $a\in M\isect A$.  
Now, $B\isect I\subseteq M$ implies $B\isect I\subseteq N$, $C\isect
M=\emptyset$ implies $C\isect N=\emptyset$, and $a\in M$ implies $a\in
N$ and $\overline{a}\not\in N$. Together with $\{A\}\IF B, \naf
C\in P$ these imply $r\in\GLred{R}{N}{I}$.  
Since $M'\not\models r$, we have $a\not\in M'$ and $B\setminus
I\subseteq M'$. 
But, since $N'\isect\hb{\module{P}}=M'$, this implies $N'\not\models
r$, a contradiction to $N'\models \GLred{R}{N}{I}$.
\item
$r$ is of the form $a\IF\limit{w'}{B\setminus
I=W_{B\setminus I}}$.
Then there is a weight rule
$a\IF\limit{w}{B=W_{B},\naf C=W_{C}}\in P$ such that
$$w'=\max(0,w-\sum_{b\in B\isect I\isect M} w_b -\sum_{c\in C\setminus
M}w_c).$$
Since $M'\not\models r$, we have $a\not\in M'$ and $w'\leq
\sum_{b\in (B\setminus I)\isect M'}w_b$. 
Define $B'=(B\isect I\isect M)\union((B\setminus I)\isect M')$ and
$C'=C\setminus M$, and recall that $N'\isect\hb{\module{P}}=M'$. 
Now $w\leq\sum_{b\in B'}w_b+\sum_{c\in C'}w_c$, 
$C'\isect N=\emptyset$, $B'\isect I\subseteq N$, $B'\setminus
I\subseteq N'$ and $a\not\in N'$, which  
implies that there is a rule $r'=a\IF B'\setminus
I\in\GLred{R}{N}{I}$ such that $N'\not\models r'$ which is in
contradiction with $N'\models\GLred{R}{N}{I}$.
\end{enumerate}
Thus assuming that there is $N'\subset N\setminus I$ such that
$N'\models \GLred{R}{N}{I}$ leads to a contradiction, and it holds
that  
$N\setminus I=\lm{\GLred{R}{N}{I}}$, that is,
$N\in\sm{\tr{NLP}{\module{P}}}$. 

\item[(ii)] \emph{We show that $f:\;\sm{\module{P}}\rightarrow
    \sm{\tr{NLP}{\module{P}}}$ is a bijection:}

Clearly $f$ is an injection: $M\ne M'$ implies $f(M)\ne f(M')$. 
To show that $f$ is a surjection, we consider an arbitrary
$N\in\sm{\tr{NLP}{\module{P}}}$ and show 
$N\isect\hb{\module{P}}=M\in\sm{\module{P}}$ and 
$f(M)=f(N\isect\hb{\module{P}})=N$.

\begin{itemize}
\item
Assume first $M\not\models\GLred{P}{M}{I}$, that is, there is a rule 
$r\in \GLred{P}{M}{I}$ that is not satisfied. 
Notice that all the rules in $\GLred{P}{M}{I}$ corresponding to a 
choice rule in $P$ are always satisfied in $M$. Thus we need to
consider only rules that correspond to a weight rule in $P$.
Now, $r=a\IF\limit{w'}{B\setminus I=W_{B\setminus
    I}}\in\GLred{P}{M}{I}$, if there is a weight rule 
$a\IF\limit{w}{B=W_{B},\naf C=W_{C}}\in P$ such 
that
$w'=\max(0, w-\sum_{b\in B\isect M\isect I} w_b-\sum_{c\in C\setminus
  M}w_c)$. 
Since $M\not\models r$, then $a\not\in M$ and $w'\leq\sum_{b\in
  (B\setminus I)\isect M}w_b$.
Define $B'=B\isect M$ and $C'=C\setminus M$.
Since $N\isect\hb{\module{P}}=M$ and
$w\leq \sum_{b\in B'}w_b+\sum_{c\in C'}w_c$, 
there is a normal rule $a\IF B',\naf C'\in R$.
Furthermore, $C'\isect N=\emptyset$ and 
$B'\isect I\subseteq N$ imply $r'=a\IF B'\setminus
I\in\GLred{R}{N}{I}$.   
Since $a\not\in M$, also $a\not\in N$. 
Furthermore, $B'\setminus I\subseteq M\subseteq N$.
These imply $N\not\models r'$ which leads to a contradiction.
Thus $M\models r$, and moreover $M\models\GLred{P}{M}{I}$. 

\item
Next, assume that there is $M'\subset M\setminus I$ such that $M'\models  
\GLred{P}{M}{I}$ and define 
$N'=M'\union(N\setminus\hb{\module{P}})\subset N\setminus I$. 
Since $N'\setminus\hb{\module{P}}=N\setminus\hb{\module{P}}$ by definition, 
each rule of the form $\overline{a}\in\GLred{R}{N}{I}$ 
is satisfied in $N'$.
Other rules in $\GLred{R}{N}{I}$ are of the form $r=a\IF B'\setminus
I$ where $a\in\hb{\module{P}}$. 
There are now two possibilities.
\begin{enumerate}
\item
There is a choice rule $\choice{A}\IF
B',\naf C\in P$, such that $a\in 
A$, $B'\isect I\subseteq N$, $C\isect N=\emptyset$, and 
$\overline{a}\not\in N$.
Now, $M\isect\hb{\module{P}}=N\isect\hb{\module{P}}$ implies 
$r\in\GLred{P}{M}{I}$, and furthermore, 
$M'\models \GLred{P}{M}{I}$ implies $M'\models r$. 
Recalling $M'\isect\hb{\module{P}}=N'\isect\hb{\module{P}}$, we get
$N'\models r$.
\item
There is a rule $a\IF B', \naf C'\in R$ corresponding to a weight rule
$a\IF\limit{w}{B=W_{B},\naf C=W_{C}}\in P$ such that   
$B'\subseteq B$, $C'\subseteq C$,
$w\leq \sum_{b\in B'}w_b+\sum_{c\in C'}w_c$, $B'\isect I\subseteq N$
and $C'\isect N=\emptyset$.
If $B'\setminus I\not\subseteq N'$, then $N'\models r$.
Assume that $B'\setminus I\subseteq N'$.
It follows from $B'\subseteq B$ and $C'\subseteq C$ that
$$w\leq\sum_{b\in (B\setminus I)\isect N'}w_b+
\sum_{b\in B\isect I\isect N}w_b+\sum_{c\in C\setminus N}w_c.$$ 
Since $M\isect\hb{\module{P}}=N\isect\hb{\module{P}}$, there is
$r'=a\IF\limit{w'}{B\setminus I=W_{B\setminus I}}\in\GLred{P}{M}{I}$
such that $w'=\max(0,w-\sum_{b\in B\isect I\isect M}w_b+\sum_{c\in
C\setminus M}w_c)$. 
Furthermore, $w'\leq\sum_{b\in (B\setminus I)\isect M'}w_b=
\sum_{b\in (B\setminus I)\isect N'}w_b$.
Since $M'\models r'$, we have $a\in M'$, and also $a\in N'$. Thus 
$N'\models r$. 
\end{enumerate}
Thus using the assumption $M\setminus I\ne\lm{\GLred{P}{M}{I}}$ we
can show that there is $N'\subset N\setminus I$ such that  
$N'\models\GLred{R}{N}{I}$, which leads to a contradiction with
$N\setminus I=\lm{\GLred{R}{N}{I}}$. Therefore, $M\setminus
I=\lm{\GLred{P}{M}{I}}$.  

\item
Finally, we show that $f(M)=f(N\isect\hb{\module{P}})=N$. 
Let $N'=f(M)$, that is,   
$$N'=
M \union\{\overline{a}\mid 
a\in\choiceheads{P}\setminus M\}.$$
Notice that $N\isect \hb{\module{P}}=N'\isect \hb{\module{P}}=M$. 
Assume $N\not\subseteq N'$, that is,  there is $\overline{a}\in N$ such that 
$\overline{a}\not\in N'$. Since $\overline{a}\not\in N'$, we have
$a\in N'$ and furthermore $a\in N$. 
The only rule $r$ in $R$ such that $\overline{a}=\head{r}$ is
$\overline{a}\IF \naf a$. However, if $a\in N$, there is no rule 
in $\GLred{R}{N}{I}$ in which $\overline{a}$ appears in the head.
Because $N\setminus I$ is the least model of $\GLred{R}{N}{I}$, we
have $\overline{a}\not\in N$, a contradiction. 
Assume then $N'\not\subseteq N$, that is, there is $\overline{a}\in N'$
such that $\overline{a}\not\in N$. Since $\overline{a}\in N'$, we have
$a\not\in N'$ and furthermore $a\not\in N$. 
If $a\not\in N$, then $\overline{a}\in\GLred{R}{N}{I}$. Since
$N\setminus I\models\GLred{R}{N}{I}$, we must have $\overline{a}\in
N$, a contradiction.  
Therefore it holds that $N=N'$.
\end{itemize}
\end{itemize}
Thus we have shown that $\trop{NLP}$ is a strongly faithful,
$\join$-preserving, and modular translation function. 
\end{proof}

%------------------------------------------------------------------------------

\end{document}